\documentclass[sigconf]{acmart}
\usepackage{booktabs}
\usepackage{cleveref}
\usepackage{wrapfig}
\usepackage{amsmath,amsfonts,bm}

\def\1{\bm{1}}

\def\rvx{{\mathbf{x}}}

\DeclareMathAlphabet{\mathsfit}{\encodingdefault}{\sfdefault}{m}{sl}
\SetMathAlphabet{\mathsfit}{bold}{\encodingdefault}{\sfdefault}{bx}{n}

\def\gG{{\mathcal{G}}}

\def\gN{{\mathcal{N}}}

\usepackage{bmpsize}

\usepackage{algorithm}
\usepackage{amsmath}
\usepackage[noend]{algpseudocode}
\usepackage{bbm}

\AtBeginDocument{%
  \providecommand\BibTeX{{%
    \normalfont B\kern-0.5em{\scshape i\kern-0.25em b}\kern-0.8em\TeX}}}

\copyrightyear{2023}
\acmYear{2023}
\setcopyright{rightsretained}
\acmConference[KDD '23]{Proceedings of the 29th ACM SIGKDD Conference on Knowledge Discovery and Data Mining}{August 6--10, 2023}{Long Beach, CA, USA}
\acmBooktitle{Proceedings of the 29th ACM SIGKDD Conference on Knowledge Discovery and Data Mining (KDD '23), August 6--10, 2023, Long Beach, CA, USA}
\acmDOI{10.1145/3580305.3599394}
\acmISBN{979-8-4007-0103-0/23/08}

\usepackage{soul}
\usepackage{amsmath}
\usepackage{bbm}

\usepackage[noend]{algpseudocode}
\begin{document}


\title{Interpretable Graph Neural Network-based Sparsification of Brain Graphs: Better Practices and Effective Designs}
\title{Interpretable Sparsification of Brain Graphs: \\Better Practices and Effective Designs for Graph Neural Networks}

% % \authornote{Both authors contributed equally to this research.}
% \affiliation{%

\author{Gaotang Li}
\affiliation{
    \institution{University of Michigan, Ann Arbor}
    \city{}
    \country{}
}
\email{gaotang@umich.edu}

\author{Marlena Duda}
\affiliation{%
  \institution{Georgia State University}
  \city{}
  \country{}
}
\email{mduda@gsu.edu}

\author{Xiang Zhang}
\affiliation{%
  \institution{University of North Carolina, Charlotte}
  \city{}
  \country{}
}
\email{xiang.zhang@uncc.edu}

\author{ Danai Koutra}
\affiliation{%
  \institution{University of Michigan, Ann Arbor}
  % \streetaddress{3223 S State St}
  \city{}
  \country{}
}
\email{dkoutra@umich.edu}

\author{ Yujun Yan}
\affiliation{%
  \institution{Dartmouth College}
  \city{}
  \country{}
}
\email{yujun.yan@dartmouth.edu}

\renewcommand{\shortauthors}{Gaotang Li, Marlena Duda, Xiang Zhang, Danai Koutra, \& Yujun Yan}
\newcommand{\reminder}[1]{\textcolor{red}{[#1]}}
\newcommand{\revision}[1]{\textcolor{black}{#1}}

\newcommand{\xiang}[1]{{\color{black}#1}}
\newcommand{\oliver}[1]{{\color{black}#1}}
\newcommand{\yy}[1]{{\color{black}#1}}

% \newcommand{\edge_mask}{\mathcal{M}}
% \newcommand{\edge_im_mask}[0]{\mathcal{M}}

% \usepackage{algpseudocode}
% \algrenewcommand\algorithmicrequire{\textbf{Input:}}
% \algrenewcommand\algorithmicensure{\textbf{Output:}}
\begin{abstract}

Brain graphs, which model the structural and functional relationships between brain regions, are crucial in neuroscientific 
% (e.g., understanding brain activities) 
and clinical applications \revision{involving graph classification}.
% (e.g., predicting neurological phenotypes and psychiatric disorders)
However, dense brain graphs pose computational challenges \revision{including high runtime and memory usage and limited interpretability.} In this paper, we investigate effective designs in Graph Neural Networks (GNNs) to sparsify brain graphs by eliminating noisy edges. %The key in graph sparsification is identifying a set of noisy edges to remove. 
% Many prior works \revision{remove} noisy edges based on explainability or task-irrelevant properties, but this does not guarantee performance improvement when using the sparsified graphs.
\revision{While prior works remove noisy edges based on explainability or task-irrelevant properties, their effectiveness in enhancing performance with sparsified graphs is not guaranteed. Moreover, existing approaches often overlook collective edge removal across multiple graphs.}
% Additionally, the selection of noisy edges is often tailored to each individual graph, making it challenging to sparsify multiple graphs collectively using the same approach.

% \textcolor{purple}{does not have a clear idea in mind treating this change}
% Additionally, the selection of noisy edges is often specific to the graph, making it unsuitable for the joint sparsification of multiple graphs.

% To address the issues above, we introduce an iterative framework to analyze the effectiveness of different sparsification models. By utilizing this framework, we find that (i) methods that prioritize interpretability may not be suitable for graph sparsification, as the sparsified graphs may degenerate the performance of GNN models;
% (ii) it is beneficial to learn the edge selection during the training of the GNN, rather than after the GNN has converged; (iii) learning a joint edge selection shared across all graphs achieves higher performance than generating separate edge selection for each graph; and (iv) gradient information, which is task-relevant, helps with edge selection. Based on these insights, we propose a new model, Interpretable Graph Sparsification (\texttt{IGS}), which improves the graph classification performance by up to $5.1\%$ with $55.0\%$ fewer edges than the original graphs. The retained edges identified by \texttt{IGS} provide neuroscientific interpretations and are supported by well-established literature. 

\revision{To address these issues, we introduce an iterative framework to analyze different sparsification models. Our findings are as follows: (i) methods prioritizing interpretability may not be suitable for graph sparsification as they can degrade GNNs' performance in graph classification tasks; (ii) simultaneously learning edge selection with GNN training is more beneficial than post-training; (iii) a shared edge selection across graphs outperforms separate selection for each graph; and (iv) task-relevant gradient information aids in edge selection. Based on these insights, we propose a new model, Interpretable Graph Sparsification (\texttt{IGS}), which enhances graph classification performance by up to 5.1\% with 55.0\% fewer edges. The retained edges identified by \texttt{IGS} provide neuroscientific interpretations and are supported by well-established literature.}

\end{abstract}

\ccsdesc[500]{Computing methodologies~Neural networks}
\ccsdesc[500]{Applied computing~Bioinformatics}

\keywords{Graph Neural Networks; Interpretability; Graph Sparsification}



\maketitle

\section{Introduction}

% The study of brain networks has captivated neuroscientists as they work to comprehend the organization of the human brain and forecast clinical outcomes. \reminder{cite} 
Understanding how brain function emerges from the communication between neural elements remains a challenge in modern neuroscience~\cite{avena2018communication}. Over the years, researchers have used brain graphs to encode the correlations of brain activities and uncover interesting connectivity patterns between brain regions. They find that the topological properties of brain graphs are useful in predicting various phenotypes and understanding brain activities~\cite{bassett2006small, bullmore2009complex, bullmore2011brain, honey2009predicting, park2008comparison}, which account for the wide usage of brain graphs in neuroscientific research~\citep{lindquist2008statistical, safavi2017scalable, yan2019groupinn}. Adopting the graph representations (often termed "connectomes"), many neuroscientific problems can be cast as graph problems. In this paper, we focus on end-to-end brain graph classification tasks since many brain graph classification tasks have meaningful real-life clinical significance, such as providing a non-invasive neuroimaging biomarker for the identification of certain psychiatric/neurological disorders at an early stage (\emph{e.g.} autism, Alzheimer's disease)~\citep{muller2014language}.

% The communication patterns between neural elements can be captured by functional brain graphs, which are obtained from Functional magnetic resonance imaging (fMRI) and encode the correlations of brain activities between the neural elements. Brain graphs have been widely used in neuroscientific research~\citep{lindquist2008statistical, safavi2017scalable}, since they 
% Among various representations of the brain, graphs 
% pre-processed 
% from functional magnetic resonance imaging (fMRI), 

% stand out as a popular choice for today's scientific methods.
% Graphs are widely used as the representative structure for the brain since they can 
% generalize well to neuroscientific data of any scale, modality, and volume while providing the potential for interpretability~\citep{bassett2006small, bullmore2009complex, bullmore2011brain}. To leverage the strengths of graph structures, 

% \reminder{Motivation for brain graphs. Why are graphs used in modeling brain activities.}

Despite the benefits of modeling brain data as graphs,
% However, compared with typical graph classification datasets, 
even well-preprocessed brain graphs pose serious challenges. A functional MRI-based (fMRI) brain graph, which is usually computed as pairwise correlations of fMRI time-series data, is fully connected. The resulting dense graph causes two unavoidable problems. 
First, it inhibits the use of efficient sparse operations, which leads to large time and memory consumption when the graphs are large~\citep{yan2019groupinn, chung2018statistical}. Second, the dense graph suffers from fMRI-related noise, making it extremely hard to train a model that learns useful generalization rules and provides good interpretability~\citep{liu2016noise}. To this end, it is crucial to make brain graphs more sparse and less noisy. The common practice in neuroscience is to remove the "weak" edges, whose weights are below the predefined threshold~\cite{power2011functional}. However, 
% apply direct thresholding, 
% but 
direct thresholding requires a wide search for the proper threshold~\cite{bordier2017graph}, and the sparsified graphs may lack useful edges and preserve significant noise. %\textcolor{red}{your framework has not been presented yet -- these results seem premature here}
To illustrate it, in Table~\ref{tab:motivation}, we show the performance on the original graphs and sparsified graphs obtained using direct thresholding %and our propsed method, respectively, 
in a classification task. It can be seen that direct thresholding may drop important edges and/or keep unimportant edges, which leads to a decrease in performance. 
Prior work related to graph sparsification generally falls into
% Related works to our problem setup 
% can be 
% generally categorized into 
two categories. The first line of work learns the relative importance of the edges, which can be used to remove unimportant edges in the graph sparsification process. These works usually
% \yy{learns an edge importance mask, a matrix that can reveal the relative importance of the edges in the graphs and can be used to sparsify the graphs.}
focus on interpretability explicitly, 
oftentimes referred to as ``explainable graph neural networks (explainable GNNs)''~\cite{yuan2022explainability}. 
The core idea embraced by this community is to identify small subgraphs that are most 
% a small subgraph of the original graph that is most 
accountable for model predictions. The relevance of the edges to the final predictions is encoded into an edge importance mask, a matrix that reveals 
% generate the edge importance mask 
% \xiang{I'm not sure, is it edge importance mask or edge importance mask?}, 
% a matrix that can reveal 
the relative importance of the edges 
% in the graphs 
and can be used to sparsify the graphs. These works show good interpretability under various measures~\cite{pope2019explainability}.
However, it remains unclear whether better interpretability indicates better performance. The other line of work \revision{tackles unsupervised graph sparsification~\citep{liu2018graph}, without employing any label information}. Some methods reduce the number of edges by approximating pairwise distances~\citep{peleg1989graph}, cuts~\citep{karger1994random}, or eigenvalues~\citep{spielman2011spectral}. \revision{These task-irrelevant methods may discard useful task-specific edges for predictions. Fewer works are task-relevant, primarily focusing on node classification~\citep{neuralSparse, PTDNet}. Consequently, these works produce different edge importance masks for each graph.} 
% As a consequence, these works generate different edge importance masks for different graphs.
\revision{However, in graph classification, individual masks can lead to significantly longer training time and susceptibility to noise. Conversely, a joint mask emerges as the preferred choice, offering robustness against noise and greater interpretability.}
% Nevertheless, since we focus on graph classification, the individual mask can contribute to significantly greater training time and may be sensitive to individual noise. Conversely, a joint mask intuitively emerges as the better choice due to its robustness against noise and the potential for \revision{more straightforward} interpretation.

\begin{table}[t!]
\caption{Brain graph classification performance (accuracy) on the original graphs (\texttt{Original}) and sparsified graphs (\texttt{Direct threholding}). \texttt{Direct thresholding} may keep unimportant edges. Details about the data and experimental setup can be found in Section~\ref{sec:exp_setup}.}
\vspace{-0.7cm}
\label{tab:motivation}
\begin{tabular}{ccc}\\\toprule  
 & PicVocab & ReadEng \\\midrule
\texttt{Original} &52.7\tiny\(\pm\)3.77 & 55.4\tiny\(\pm\)3.51\\  
\texttt{Direct thresholding} &52.0\tiny\(\pm\)5.51 & 54.8\tiny\(\pm\)3.19\\  \bottomrule
% \texttt{IGS (our method)} &57.8\tiny\(\pm\)3.10 & 60.1\tiny\(\pm\)2.78\\  \bottomrule
\end{tabular}
\vspace{-0.4cm}
\end{table}

% \reminder{Brief introduction of the literature into different categories: (1) models aimed at interpretability. They can generate the edge importance mask, a matrix which can reveal the relative importance of the edges in the graphs. This edge importance mask can be used to sparsify the graphs. (2)
% models aimed at sparsification. task irrelevant. Some are task-relevant but focus on the node classification task, so they generate different mask for different graphs.
% Current limitations: 1. Focuses on interpretability but it remains unclear whether better interpretability indicates better performance. 2. prior work focuses on the node classification task and generate different mask for different graphs. Emphasize the importance of the graph classification task in our setting, and why we need a common mask (more robust to noise, better interpretation). 3. Task irrelevant}

% \xiang{You mentioned three problems in this para, it's great. But which are the three problems are not super clear in previous paragraphs. Maybe we can use conjunctions like "First, Second, Third" to emphasize it. }
% To address the first and second problems, 
\vspace{0.1cm}
\noindent \textbf{This work.} To assess the quality of the sparsified graphs obtained from interpretable models in the graph classification task,
we propose to evaluate the effectiveness of the sparsification algorithms under an iterative framework. At each iteration, the sparsification algorithms decide which edges to remove and feed the sparsified graphs to the next iteration. We measure the effectiveness of a sparsification algorithm by computing the accuracy of the downstream graph classification task at each iteration. An effective sparsification algorithm should acquire the ability to identify and remove noisy edges, resulting in a performance boost in the graph classification task after several iterations (\Cref{subsec: iterative_framework}). %The intuition is that an effective and interpretable algorithm should acquire the ability to identify and remove noisy edges, which in turn improves performance~(\Cref{subsec: iterative_framework}).
% \xiang{Agree with the intuition. But several questions arise immediately: 1. how does the model know the performance change (increase or decrease) several iterations later? 2) what will the model do if the performance doesn't increase? Add the pruned edge back? 3) in the training of DL, performance fluctuation is common, which means the performance may increase and decrease in a short window (e.g., within 5 iterations). We believe the DL model is healthy as long as the trend of performance (e.g., in 10 iterations) in increasing. How does the proposed model deal with the fluctuation?}

We utilize this iterative framework to evaluate two common practices used in graph sparsification and graph explainability\revision{: (1)} obtaining the edge importance mask from a trained model and \revision{(2)} learning an edge importance mask for each graph individually~\cite{yuan2022explainability}. For instance, \texttt{GNNExplainer}~\cite{ying2019gnnexplainer} learns a separate edge importance mask for each graph \textbf{after} the model is trained. 
% \textcolor{blue}{Xiang: can we add an appendix to provide the experimental setup and results of the "empirical analysis"? }
Through our empirical analysis, we find that these practices are \textbf{not helpful} in graph sparsification, as the sparsified graphs may lead to lower classification accuracy. \textcolor{black}{In contrast, we identify three key strategies that can improve the performance.} 
Specifically, we find that \textbf{(S1)} learning a \ul{\textbf{joint} edge importance mask 
\textbf{(S2)} \textbf{simultaneously with the training} of the model} helps improve the performance over the iterations, as it passes task-relevant information through back-propagation. 
% Another advantage of this strategy is %training the mask with the model is 
% that task-relevant information will be passed through back-propagation. 
Another strategy to incorporate the task-relevant information is to \textbf{(S3)} \ul{initialize the mask with the gradient information from the immediate previous iteration}. This strategy is inspired by the evidence in the computer vision domain that gradient information may encode data and task-relevant information and may contribute to the explainability of the model~\cite{adebayo2018sanity, hong2015online, alqaraawi2020evaluating}. 
% We combine these strategies and 

Based on the identified strategies, we propose a new \textbf{I}nterpretable model for brain \textbf{G}raph \textbf{S}parsification, \texttt{IGS}. We evaluate our \texttt{IGS} model on real-world brain graphs under the iterative framework and find that it can benefit from iterative sparsification. 
% Compared to the competitive baselines \oliver{maybe delete this compared to?}, 
\texttt{IGS} achieves up to 5.1\% improvement on graph classification tasks with graphs of 55.0\% fewer edges than the original compared to strong baselines. %Thanks to its careful selection of important edges, it also \textcolor{red}{significantly outperforms the common practice in neuroscience of directly thresholding graphs in order to sparsify them~\cite{??} (more details about the data and experimental setup can be found in Section~\ref{sec:exp_setup}).}. These results demonstrate that thresholding can indeed drop important edges and retain noisy ones.

\begin{figure*}[!th]
    \centering
    \includegraphics[width=\textwidth]{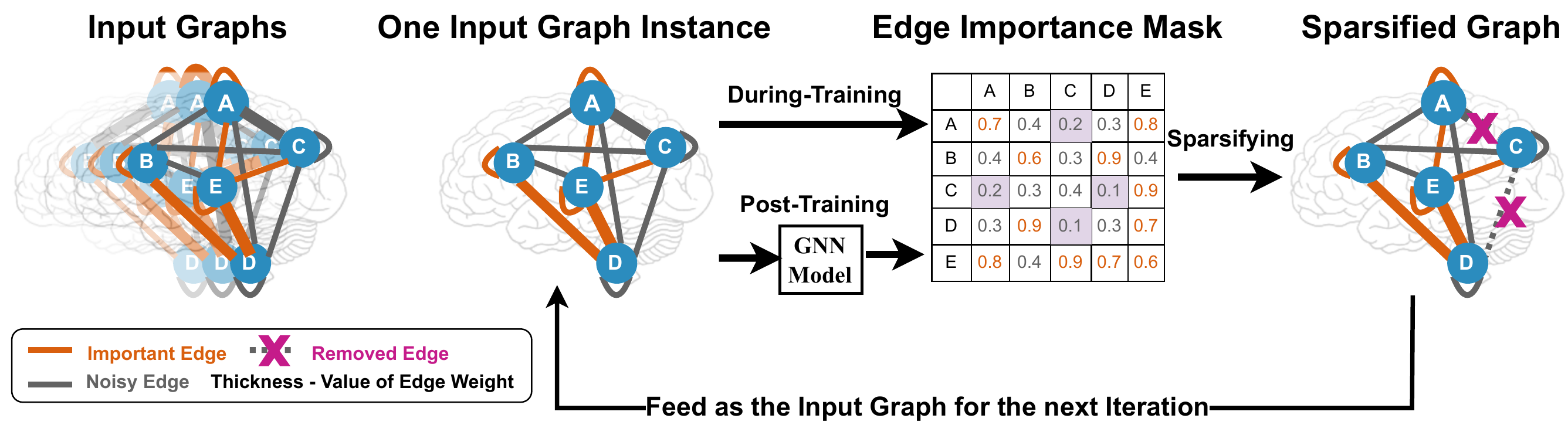}
    \caption{General iterative framework of sparsification. This framework progressively eliminates noisy edges from input brain graphs by learning an edge importance mask for each/all graph(s). The edge importance mask(s) can be generated from a well-trained GNN model or trained simultaneously with a GNN model. Important edges are depicted in orange, while noisy edges are shown in grey. Dashed lines with purple crosses represent the removed edges in the sparsified graphs. }
    \label{fig:intro_fig}
    % \vspace{-3mm}
\end{figure*}

Our main contributions are summarized as follows:
\vspace{-2mm}
\begin{itemize}
    \item \textbf{General framework.} We propose a general iterative framework to analyze 
    % improve the evaluation of
    % \reminder{evaluate} 
    the effectiveness of different graph sparsification models.
    We find that edge importance masks generated from interpretable models may not be suitable for graph sparsification because they may not improve the performance of graph classification tasks.
    \item \textbf{New insights.} We find that two practices commonly used in graph sparsificati\revision{o}n and graph explainability are not helpful under the iterative framework. Instead, we find that learning a joint edge importance mask along with the training of the model improves the classification performance \revision{during} iterative graph sparsification. \yy{Furthermore, incorporating gradient information in mask learning \revision{also} boosts the performance in iterative sparsification.}
    \item \textbf{Effective model.} Based on the insights, we propose a new model, \texttt{IGS}, which can improve the performance (up to 5.1\%) with \revision{significantly} sparser graphs (up to 55.0\% less edges).
    \item \textbf{Interpretability.} Our \texttt{IGS} model learns to remove task-irrelevant edges in the iterative process. The edges that are retained by \texttt{IGS} have neuroscientific interpretations and are well supported by well-established literature.
\end{itemize}

\section{Notation and Preliminaries}
\label{sec:prelim}
% In this section, we introduce the necessary preliminaries to understand this paper.
% % the problem formulation 
% % in this paper. 
% We begin with the \revision{key} notations. Then, we briefly explain the background of graph neural networks. Finally, we formally define the problems \revision{that we investigate}.
\revision{In this section, we introduce key notations, provide a brief background on GNNs, and formally define the problem that we investigate.}

\vspace{0.1cm}
\noindent \textbf{Notations.} 
We consider a set of graphs \(\mathcal{G}\). Each graph $G_i(\mathcal{V}, \mathcal{E}_i) \in \mathcal{G}$ in this set has $n$ nodes, and the corresponding node set and edge set are denoted as $\mathcal{V}$ and \revision{$\mathcal{E}_i$}, respectively. \xiang{The graphs share the same set of nodes.}
% Let the set of input graphs be denoted as \(\mathcal{G}\), consisting of a pair of node and edge set \((\mathcal{V}, \mathcal{E})\) respectively. Suppose each input graph $G \in \mathcal{G}$ has \(n\) nodes, then 
The set of neighboring nodes of node \revision{$v$} is denoted as \revision{$\gN_v$}.
We focus on the setting where the input graphs are weighted, and we represent the weighted adjacency matrix of each input graph \(G_i\) as \(\mathbf{A}_i \in \mathbb{R}^{n \times n}. \) The node features in $G_i$ are represented by a matrix \(\mathbf{X}_i \in \mathbb{R}^{n \times d}\), where its $j$-th row $\mathbf{X}_i{[j,:]}$ represents the features of the $j$-th node, and \(d\) refers to the dimensionality of the node features. For conciseness, we use $\mathbf{X}_i^{(l)}$ to represent the node representations/output at the $l$-th layer of a GNN. 
% , where \(d_e\) refers to the dimensionaly of edge features. 
Given our emphasis on graph classification problems, we denote the number of classes as $k$, the set of labels as 
 \(\mathcal{Y}\), and associate each graph $G_i$ with a corresponding label $y_i \in \mathcal{Y}$.
% Since we focus on graph classification problems, 
% % we use \(\mathcal{Y} \in \mathbb{R}^{k}\) to represent the set of labels, where \(k\) refers to the number of classification classes. Each graph $G_i$ is associated with label $y_i \in \mathcal{Y}$. 
% we use $k$ to represent the number of classes, \(\mathcal{Y}\) to represent the set of labels, and each graph $G_i$ is associated with label $y_i \in \mathcal{Y}$.

\revision{We also} leverage gradient information\revision{~\citep{simonyan2013deep}} in this work\revision{:} 
% We present the notations as follows: 
% we use \(\nabla f_j (G_i)\) to represent the gradients with respect to the input graph $G_i$, which are backpropagated from taking the output in class $j$. We call it the gradient map. 
\(\nabla f_j (G_i)\) denote\revision{s} the gradients of the output in class $j$
with respect to the input graph $G_i$. These gradients are obtained through backpropagation and are referred to as the gradient map.
% $\mathbf{T}^{(j)}_i=\nabla f_j (G_i)$.} 
% We assume the softmax layer is applied at the end of the model architecture. 

% \xiang{We need to further polish the notation, but can leave as it is for now.}

\vspace{0.1cm}
\noindent \sloppy \textbf{Supervised Graph Classifcation.} Given a set of graphs $\{G_1, G_2, \cdots, G_t\}$ and their labels $\{y_1, y_2, \cdots, y_t\}$ for training, we aim to learn a function $f \colon \mathcal{G} \to \mathcal{Y}$,  such that the loss $\mathbb{E}(\mathcal{L}(y_i, \hat{y_i}))$ is minimized, where $\mathbb{E}$ denotes expectation, $\mathcal{L}$ denotes a loss function, and $\hat{y_i}=f(G_i)$ denotes the predicted label of $G_i$.

\vspace{0.1cm}
% \textcolor{purple}{zhe li yun le}

\noindent \textbf{GNNs for Graph Classification.} An $L$-layer GNN model~\cite{kipf2017semi, velickovic2018graph, xu2018powerful, yan2019groupinn, yan2021two} often follows the message-passing framework, which consists of three components~\cite{gilmer2017neural}: (1) neighborhood propagation and aggregation: \revision{\small $\mathbf{m}_v^{(l)}=\texttt{AGGREGATE}(\mathbf{X}_i^{(l)}[u,:]$, $u \in \gN_v$)}; (2) combination: \revision{\small $\mathbf{X}_i^{(l+1)}[v,:]$ = \texttt{COMBINE}($\mathbf{X}_i^{(l)}[v,:]$, $\mathbf{m}_v^{(l)}$)}, where {\small \texttt{AGGREGATE}} and {\small \texttt{COMBINE}} are learnable functions; (3) global pooling. \revision{$\rvx^{G_i}$ = \texttt{Pooling} ($\mathbf{X}_i^\text{(L)}\}$)}, {where the \texttt{Pooling} function operates on all node representations, including options like \texttt{Global\_mean}, \texttt{Global\_max} or other complex pooling functions~\cite{ying2018hierarchical, knyazev2019understanding}.} The loss is given by $L$ = $\frac{1}{N^{G}}\sum_{G_i \in \mathcal{G}_{\text{train}}}$\texttt{CrossEntropy} (\texttt{Softmax}($\rvx^{G_i}$), $y_i$), where $\gG_{\text{train}}$ represents the set of training graphs and $N^{G}=|\gG_{\text{train}}|$. Though our framework does not rely on specific GNNs, we illustrate the effectiveness of our framework using the GCN model proposed in~\cite{kipf2017semi}.

% The message passing and aggregation mechanisms can be considered together as the propogation step. Specifically, at the \(l\)-th GNN layer, it first computes the message between a node pair \((v_i, v_j)\) based on their representations \(h_i^{l-1}, h_j^{l-1} \) in previous layer. Then, for each node \(v_i\), an aggregator merges the messages passed from its local neighborhood. Finally, each node will be updated with a new representation based on its previous representations and the output from the aggregator. 

The performance of GNN models heavily depends on the quality of the input graphs. Messages propagated through noisy edges can significantly affect the quality of the learned representations~\cite{yan2019groupinn}. Inspired by this observation, we focus on the following problem:

\vspace{0.1cm}
\noindent \textbf{Problem: \revision{Interpretable}, Task-relevant Graph Sparsification.}

Given a set of input graphs $\mathcal{G}=\{G_1, G_2, \cdots, G_t\}$ and the corresponding labels $\mathcal{Y}=\{y_1, y_2, \cdots, y_t\}$, we \revision{seek to learn} a set of \revision{graph-specific} edge importance masks $\{\mathcal{M}_1, \mathcal{M}_2, \cdots, \mathcal{M}_t\} \in \{0,1\}^{n \times n}$,
\textbf{OR} a joint edge importance mask \(\mathcal{M}\in \{0,1\}^{n \times n}\) shared by all graphs, which can be used to remove the noisy edges and retain the most task-relevant ones. \revision{This should lead to enhanced classification performance on sparsified graphs.}
Edge masks that effectively identify task-relevant edges are considered to be interpretable.
% \textit{For an input graph \(G = (V, E)\), we aim at obtaining a subgraph \(G' = (V, E')\) of it that removes the noisy and keeps the most task-relevant edges. We hope to obtain $G'$ from $G$ through learning an indicator edge  
% \(\mathcal{M}\), where \(\mathcal{M} = \{(0,1)\}^{n \times n \times d_e}\).  Ideally, with the exactly same GNN model \(f\), \(f(G')\) should yield a better performance than \(f(G)\). Moreover, \(G\) should be interpretable. }
% \xiang{1. in the last sentence, I guess you mean G', or G and G'? 2. what exactly does interpretable means?}
\section{Proposed Method: IGS}
% In this section, we first introduce the iterative framework that we use to evaluate the effectiveness of the sparsification algorithms. Next, we elaborate three strategies used in IGS.
% present how we train the edge importance mask along with any GNN model in the graph classification task.
% In this section, we first introduce the iterative framework used to evaluate the effectiveness of the sparsification algorithms. Next, we elaborate on the strategies employed in \revision{our proposed algorithm, } \texttt{IGS} \revision{(Interpretable Graph Sparsification)}.
\revision{In this section, we introduce our proposed iterative framework for evaluating various sparsification methods. Furthermore, we introduce IGS, a novel and interpretable graph sparsification approach that incorporates three key strategies: (S1) joint mask learning, (S2) simultaneous learning with the GNN model, and (S3) utilization of gradient information. We provide detailed explanations of these strategies in the following subsections.}

\begin{figure*}[!th]
    \centering
    \includegraphics[width=0.9\textwidth]{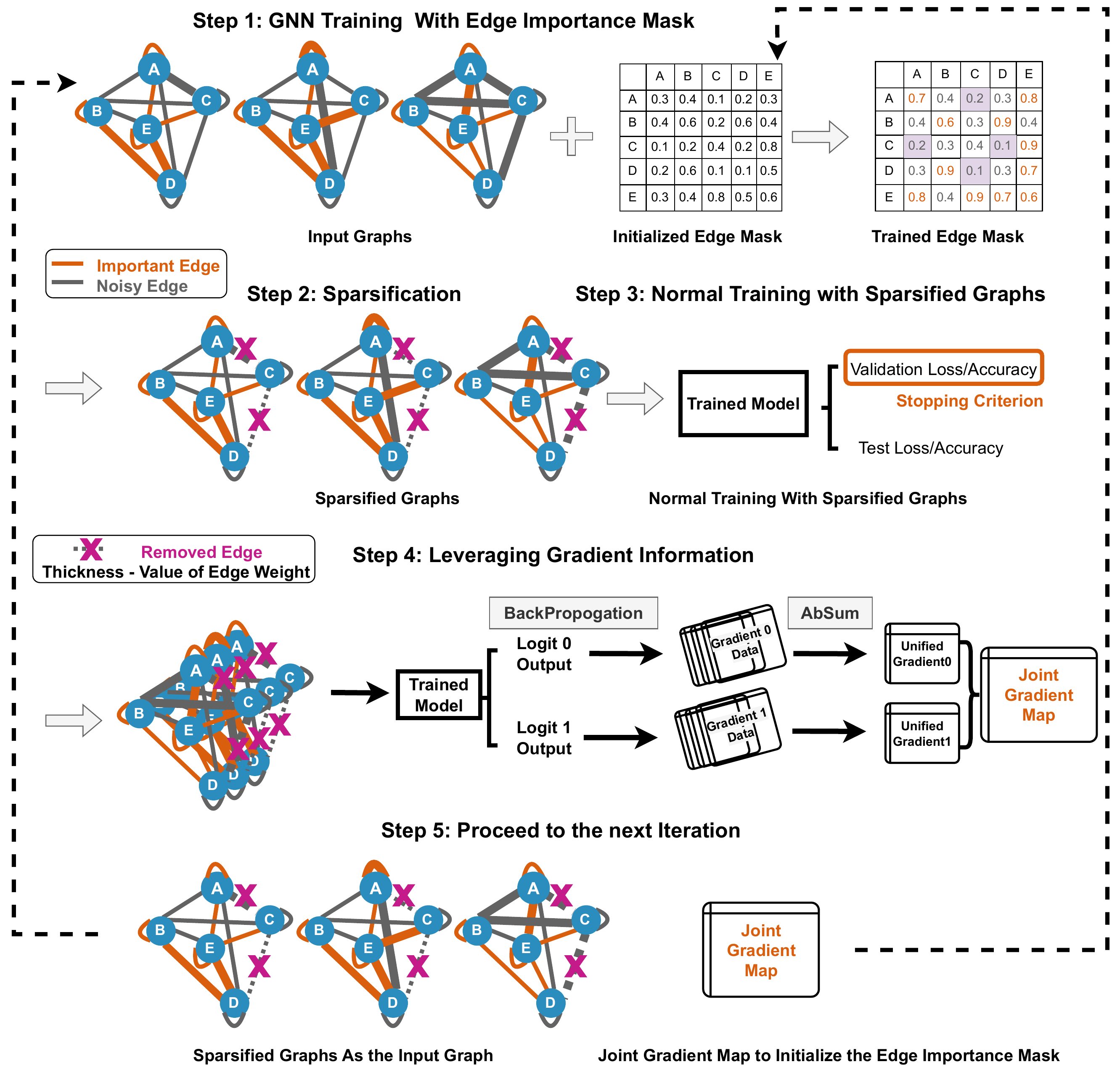}
    \caption{Training process of \texttt{IGS}. At iteration $i$, \texttt{IGS} takes a set of input graphs and initializes its joint edge importance mask using the joint gradient map from the previous iteration. It trains the GNN model and the edge importance mask together, followed by sparsifying all input graphs using the obtained mask. Normal training is then conducted on the sparsified graphs. The gradient information is later extracted by computing a joint gradient map.
    Finally, \texttt{IGS} feeds the sparsified graphs to the next iteration and uses the joint gradient map to initialize the subsequent joint edge importance mask. \xiang{\texttt{IGS} is model-agnostic and can be seamlessly integrated with existing GNN models.}
    }
    \label{fig:trained_mask} 
    \vspace{-3mm}
\end{figure*}

% \subsection{Problem Formulation}
\subsection{Iterative Framework}
\Cref{fig:intro_fig} illustrates the general iterative framework. 
At a high level, \revision{given a sparsification method, our} framework iteratively removes unimportant edges based on the edge importance masks generated by the method at each iteration.
In detail, the method can generate either \yy{a separate edge importance mask $\mathcal{M}_i$ for each input graph $G_i$ or a joint edge importance mask $\mathcal{M}$ shared by all input graphs $\mathcal{G}=\{G_1, G_2, \cdots\}$. These edge importance masks indicate the relevance of edges to the task's labels.
% Most existing works~\cite{cui2021brainnnexplainer, ying2019gnnexplainer, luo2020parameterized} generate edge importance masks from a well-trained model, where mask learning and model training are separate processes. 
In our setting, we also allow training the masks simultaneously with the model. 
Ideal edge masks are binary, where zeros represent unimportant edges to be removed. In reality, many \revision{models (\emph{e.g.} GNNs~\cite{ying2019gnnexplainer, zhang2021relex})} learn soft edge importance masks with values between [0,1]. 
% Despite this, we can still filter out a fixed percentage of edges whose importance scores are the lowest.
\revision{In each iteration, our framework removes either the edges with zero values in the masks (if binary) or a fixed percentage $p$ of edges with the lowest importance scores in the masks.} \revision{We present the framework of iterative sparsification in Algorithm~\ref{alg:iterative_sparsification},  where $\mathcal{G}^i$ denotes {the set of sparsified graphs at iteration $i$, and $G_j^i$ denotes the $j$-th graph in the set $\mathcal{G}^i$.}}
% In our framework, we set this removing percentage $p$ to be 5\% based on the cross-validation results. 

Though existing works~\cite{hooker2019benchmark, pope2019explainability} have proposed different ways to define the "importance" of an edge and thus they generate different sparse graphs, \textit{we believe that a direct and effective way to evaluate these methods is to track the performance of these sparsified graphs} under this iterative framework. The trend of the performance reveals the relevance of the remaining edges to the predicted labels.}
% an edge importance mask, indicating the relative importance of the corresponding connection. 
% Notably, according to existing works and our proposed method, the edge importance mask is either generated after training or along with training. 
% Regardless of the approach, the edge importance mask can be used to sparsify the input graph through a small thresholding. 
% Intuitively, by gradually pruning the input graph, methods can stably remove noise and achieve performance improvements. Under our framework, there will be a required hyperparameter: prune threshold $T$. 
%
% Below we present the general algorithm 

% \oliver{need help with the package issue; cannot use "Input" for some reason} \\ 

% \begin{algorithm}
% \caption{Iterative Sparsification}
% \label{alg:iterative_sparsification}
% \hspace*{\algorithmicindent} \textbf{Input:} Sparsification Method $S$, Input Graph Set $\mathcal{G}$, Training Set Index $I_{\text{train}}$\\
% \hspace*{\algorithmicindent} \textbf{Output} 
% \begin{algorithmic}[1]

% \end{algorithmic}    
% \end{algorithm}
% \begin{algorithm}\label{algo:ProclivityProp2}
% \caption{Iterative Sparsification}
% \begin{algorithmic}
%   \Input
%   % \Desc{T}{matrix of measurements}
%   \yy{Sparsification Method $S$}, 
%         % Training Method Hyperparameter $H$, 
%         Input Graph Set $\mathcal{G}$, \yy{Training Set Index} $I_{\text{train}}$, Number of Iterations $n$, GNN model $f$\\
%   \EndInput
%   \Output
%   \Desc{$\hat X$}{matrix of graph signals}
%   \Desc{$\hat W$}{matrix of outliers}
%   \EndOutput
% \end{algorithmic}
% \end{algorithm}

\begin{algorithm}
  \caption{Iterative Sparsification Framework}
  \label{alg:iterative_sparsification}
  \begin{flushleft}
        \textbf{INPUT:} \yy{Sparsification Method $S$, 
        % Training Method Hyperparameter $H$, 
        Input Graph Set $\mathcal{G}^{1}$, Graph Labels $\mathcal{Y}$, Training Set Index $\mathbbm{1}_{\text{Train}}$, Validation Set Index $\mathbbm{1}_{\text{Val}}$, Number of Iterations $N$, a GNN model}\\
\end{flushleft}
   % \hspace*{\algorithmicindent} \textbf{Input} \\
   %  \hspace*{\algorithmicindent} \textbf{Output}
  \begin{algorithmic}[1]
    % \Procedure{Iterative Sparsification}{}
    \For {i = 1, \ldots, N}
    \If {$S.$ \text{MaskTime ( ) == }\emph{PostTrain}}
    \State {\text{GNN\_Trained} $\gets$ \text{Train (GNN, } $\mathcal{G}^{i}[\mathbbm{1}_{\text{Train}}]$, $\mathcal{Y}[\mathbbm{1}_{\text{Train}}]$ \text{)}}
    \If {$S.$ \text{MaskType ( ) == }\emph{Individual}}
    \State // {\textit{Individual Edge Importance Mask: }} 
    \State {$\mathcal{M}_j=S.$~\text{MaskTrain (GNN\_Trained, $G_j^i$, $y_i$)}}
    \State {$G_j^{i+1} \gets \mathcal{M}_j \odot G_j^{i}, \ \forall j$}
    \Else
    \State // {\textit{Joint Edge Importance Mask: }}
    \State {$\mathcal{M}=S.$~\text{MaskTrain (GNN\_Trained, $\mathcal{G}^{i}$, $\mathcal{Y}$, $\mathbbm{1}_{\text{Train}}$)}}
    \State {$G_j^{i+1} \gets \mathcal{M} \odot G_j^{i}, \ \forall j$}
    \EndIf
    \Else
    \State // {\textit{Joint Edge Importance Mask: }} 
    \State {$\mathcal{M}=S.$~\text{MaskTrain (GNN, $\mathcal{G}^{i}$, $\mathcal{Y}$, $\mathbbm{1}_{\text{Train}}$)}}
    \State {$G_j^{i+1} \gets \mathcal{M} \odot G_j^{i}, \ \forall j$}
    \EndIf
    \State {\text{Validation loss $L^{i}=$ Train\&Val (GNN,}}
    \State {\text{$\mathcal{G}^{i+1}$, $\mathcal{Y}$, $\mathbbm{1}_{\text{Train}}$, $\mathbbm{1}_{\text{Val}}$)}}
    \EndFor
  \end{algorithmic}
  
\begin{flushleft}
        \textbf{OUTPUT:} $\mathcal{G}^{i}$ with smallest $L^{i}$\\
\end{flushleft}
\end{algorithm}

% \reminder{We need a pic here!}
\subsection{Strategies}
\subsubsection{Trained Mask (S1+S2)}
\yy{We aim to learn a joint edge importance mask $\mathcal{M} \in 
\{0,1\}^{n \times n}$ along with the training of a GNN model, \revision{as shown in Figure~\ref{fig:trained_mask}}. Each entry in $\mathcal{M}$ represents if the corresponding edge in the original input graph should be kept (value 1) or not (value 0).
Directly learning the discrete edge mask is hard as it cannot generate gradients to propagate back. Thus, at each iteration, we learn a soft version of $\mathcal{M}$, where each entry is within $[0,1]$ and reflects the relative importance of each edge. Considering the symmetric nature of the adjacency matrix for \xiang{undirected} brain graphs, we require the learned edge importance mask to be symmetric. 
We design the soft edge importance mask as $\sigma({\Phi}^T+\Phi)$, where $\Phi$ is a matrix to be learned and $\sigma$ is the \texttt{Sigmoid} function.
A good initialization of $\Phi$ can boost the performance and accelerate the training speed. Thus, we initialize this matrix with the gradient map (\Cref{subsec: global_grad}) from the previous iteration (Step 5 in Figure~\ref{fig:trained_mask}). Furthermore, following \cite{ying2019gnnexplainer}, we regularize the training of $\Phi$ by requiring $\sigma({\Phi}^T+\Phi)$ to be sparse. Thus we apply a $l_1$ regularization on $\sigma({\Phi}^T+\Phi)$. }
In summary, we have the following training objective:
\begin{equation}
\label{eq:train_objective}
    \min \mathcal{L}( f(\mathbf{A} \odot \sigma({\Phi}^T+\Phi), \mathbf{X}), \mathcal{Y} ) + \lambda\sum_{ij} \sigma({\Phi}^T+\Phi)_{ij}
\end{equation}
where $\odot$ denotes the Hadamard product; $\mathcal{L}$ is the \texttt{Cross-Entropy} loss;
$\lambda$ is the regularization coefficient. 
% Note that we optimize the joint mask $\Phi$ across all training samples in a batch-training fashion, as our objective is to learn a shared mask.
% since we aim at learning a joint mask, the same $\Phi$ will be optimized across all training samples in a batch-training fashion in practical implementation. 
% The final indicator matrix $\mathcal{M}$ can be obtained 
% \yy{
% Then we transform this soft mask into an indicator matrix by setting the lowest $p$ percentage values to zeros:
% % . It can be expressed as:
% }
We optimize the joint mask across all training samples in a batch-training fashion to achieve our objective of learning a shared mask. Subsequently, we convert this soft mask into an indicator matrix by assigning zero values to the lowest $p$ percentage of elements:
% , we take the hard thresholding

\begin{equation}
\label{eq:threshold}
 \mathcal{M}{[i, j]} = 
    \begin{cases}
      0 & \text{if } \sigma({\Phi}^T+\Phi)_{ij} \text{ in lowest $p\%$}  \\
      1 & \text{otherwise}
    \end{cases}       
\end{equation}
% through hard thresholding of the matrix $\sigma({\Phi}^T+\Phi)$ and it 
The indicator matrix $\mathcal{M}$ can then be used to sparsify the input graph through an element-wise multiplication, \emph{e.g.} $G_i' = \mathcal{M} \odot G_i$.

\subsubsection{Joint Gradient Information (S3)}
\label{subsec: global_grad}
Inspired from the evidence in the computer vision domain that gradient information may encode data and task-relevant information and may contribute to the explainability of the model~\cite{adebayo2018sanity, hong2015online, alqaraawi2020evaluating}, we utilize the gradient information, i.e., gradient maps to initialize and guide the learning of the edge importance mask. 

% \revision{More generally, for how we generate a joint gradient map that combines the gradient information from each training graph in the case of two classes, step 4}}
% % We propose a method to generate a global edge importance mask 
% % that combines individual information. 
%  in Figure~\ref{fig:trained_mask} illustrates the general idea. 
 \revision{Step 4 in Figure~\ref{fig:trained_mask} illustrates the general idea of generating a joint gradient map by combining gradient information from each training graph.}
 Each training graph $G_i$ has $k$ gradient maps $\nabla f_j(G_i), j=1, 2, \cdots, k$, \yy{each corresponding to the output in class $j$ (\Cref{sec:prelim}).}
 Instead of using the ``saliency maps''~\citep{simonyan2013deep}, which consider only the gradient maps from the predicted class, we \revision{leverage all the} gradient maps \revision{as they} provide meaningful knowledge. For \(G_1, \ldots, G_n \in \mathcal{G}_{\text{train}}\), we compute the unified mask of class 
j as the sum of the absolute values of each gradient map, represented as 
\begin{equation}
    \bigcup f_j = \sum_{i=1}^t \lvert \nabla f_j(G_i)\rvert
\end{equation}
By summing the unified masks of all classes, we generate the joint edge gradient map denoted as \(\mathbf{T} = \sum_{j=1}^k \bigcup f_j\). 
% \yy{Then we transform this joint gradient map into an indicator matrix by setting the lowest $p$ percentage values to zeros. It can be expressed as:}
% % , we take the hard thresholding

% \begin{equation*}
%  \mathbf{T}{[i, j]} = 
%     \begin{cases}
%       0 & \text{if } \mathbf{T}{[i, j]} \text{ in lowest $p\%$}  \\
%       1 & \text{otherwise}
%     \end{cases}       
% \end{equation*}
%
% to generate the final indicator edge mask. 

% \subsubsection{Formulation} 
% \newline 

\subsubsection{Algorithm}

\revision{We incorporate these three strategies into \texttt{IGS} and outline our method in Algorithm~\ref{alg:igs}:} 

\begin{algorithm}
  \caption{Interpretable Graph Sparsification: \texttt{IGS}}
  \label{alg:igs}
  \begin{flushleft}
        \textbf{INPUT:} Input Graph Dataset $\mathcal{G}^1$, Training Set Index $\mathbbm{1}_{\text{Train}}$, Validation Set Index $\mathbbm{1}_{\text{Val}}$, Removing Percentage $p$, Number of Iterations $N$, GNN model, Regularization Coeffient $\lambda$\\
\end{flushleft}
   % \hspace*{\algorithmicindent} \textbf{Input} \\
   %  \hspace*{\algorithmicindent} \textbf{Output}
  \begin{algorithmic}[0]
    \For {i = 1, \ldots, N}

    \State // \textbf{Step 1: GNN Training with Edge Importance Mask}
    \If {$i == 1$}
    \State    Initialize $\Phi$ using Xavier normal initiation.
    \Else
    \State  Initialize $\Phi$ using the previous joint gradient map $\mathbf{T}^{(i)}$
    \EndIf
    \State $\sigma({\Phi}^T+\Phi)$ $\gets$ Train (GNN, $\mathcal{G}^{i}$, $\mathcal{Y}$, $\mathbbm{1}_{\text{train}}$, $\lambda$).
    (\Cref{eq:train_objective})
    \State Obtain \textit{joint Edge Importance Mask} $\mathcal{M}$ following \Cref{eq:threshold}
    \State // \textbf{Step 2: Sparsification}
    \State $  G_j^{i+1} \gets \mathcal{M} \odot G_j^i, \ \forall j$
    % \State $\textit{Save} \ \mathcal{G}^{i}$
    % \State TrainVal()
    % $\text{Train } f(\mathcal{G}^i) \text{, get validation loss }  L^{i} \text{, joint gradient map } \mathbf{T}^{(i)}$
    \State // \textbf{Step 3: Normal Training with Sparsified Graphs}
    \State \text{Validation loss $L^{i}$, GNN\_Trained =  Train\&Val (GNN,}
    \State {\text{$\mathcal{G}^{i+1}$,
    $\mathcal{Y}$, $\mathbbm{1}_{\text{Train}}$, $\mathbbm{1}_{\text{Val}}$)}}
    \State  // \textbf{Step 4: Leveraging Gradient Information}
    \State $\mathbf{T}^{(i+1)} \gets \text{JointGradient}(\text{GNN\_Trained}, \ \mathcal{G}^{i+1}, \mathbbm{1}_{\text{Train}})$
    % \State \textbf{Step 5: Proceed to the next Iteration}
    
    \EndFor
  \end{algorithmic}
  
  \begin{flushleft}
        \textbf{OUTPUT:} $\mathcal{G}^i$ with smallest $L^{i}$ \\
\end{flushleft}
\end{algorithm}

\section{Empirical Analysis}
% \textcolor{purple}{do we need to change the questions as suggested? i.e., Strategy x}

In this section, we aim to answer the following \revision{research} questions \revision{using our iterative framework}: 
(Q1) Is learning a joint edge importance mask better than learning a separate mask for each graph?
(Q2) Does simultaneous training of the edge importance mask with the model yield better performance than training the mask \revision{separately} from the trained model? (Q3) Does the gradient information help with graph sparsification? (Q4) Is our method \texttt{IGS} interpretable? 

\subsection{Setup}
\label{sec:exp_setup}
\subsubsection{Dataset}

 We use the WU-Minn Human Connectome Project (HCP) 1200 Subjects Data Release as our benchmark dataset to evaluate our method and baselines~\citep{HCP_dataset}. \yy{The pre-processed brain graphs can be obtained from ConnectomeDB~\citep{ConnectomeDB}. These brain graphs are derived from the resting-state functional magnetic resonance imaging (rs-fMRI) of 812 subjects, where no explicit task is being performed. Predictions using rs-fMRI
 are generally harder than task-based fMRI~\cite{lv2018resting}.
 The obtained brain graphs are fully connected, and the edge weights are computed from the correlation of the rs-fMRI time series between each pair of brain regions~\cite{smith2015positive}.
 The parcellation of the brain is completed using Group-ICA with 100 components~\citep{griffanti2014ica, glasser2016multi, robinson2014msm, beckmann2004probabilistic,glasser2013minimal, fischl2012freesurfer}, which results in 100 brain regions comprising the nodes of our brain graphs. Additionally, a set of cognitive assessments were performed on each subject, which we utilized as cognitive labels in our prediction tasks. Specifically, we utilize the scores from the following cognitive domains as our labels, which incorporate age adjustment~\citep{ConnectomeDB}: }
 \begin{itemize}
     \item PicVocab (Picture Vocabulary) assesses language/vocabulary comprehension. The respondent is presented with an audio recording of a word and four photographic images on the computer screen and is asked to select the picture that most closely matches the word's meaning.
     \item ReadEng (Oral Reading Recognition) assesses language/reading decoding. The participant is asked to read and pronounce letters and words as accurately as possible. The test administrator scores them as right or wrong.
     \item PicSeq (Picture Sequence Memory) assesses the Open of episodic memory. It involves recalling an increasingly lengthy series of illustrated objects and activities presented in a particular order on the computer screen.
     \item ListSort (List Sorting) assesses working memory and requires the participant to sequence different visually- and orally-presented stimuli.
     \item CardSort (Dimensional Change Card Sort) assesses the cognitive flexibility. Participants are asked to match a series of bivalent test pictures (e.g., yellow balls and blue trucks) to the target pictures, according to color or shape. Scoring is based on a combination of accuracy and reaction time.
     \item Flanker (Flanker Task) measures a participant's attention and inhibitory control. The test requires the participant to focus on a given stimulus while inhibiting attention to stimuli flanking it. Scoring is based on a combination of accuracy and reaction time.
 \end{itemize}
 More details can be found in ConnectomeDB~\citep{ConnectomeDB}. These scores are continuous. In order to use them for graph classification, we assign the subjects achieving scores in the top third  to the first class and the ones in the bottom third to the second class.

\begin{table*}[!t]
\center 
\caption{Results of test accuracies of different approaches evaluated on six prediction tasks (PicVocab, ReadEng, PicSeq, ListSort, CardSort, and Flanker) across four data splits generated by different random seeds. We report the mean and standard deviation for each of them. The first row denotes the performance using the original graph trained by GCN~\cite{gcn}; the last column denotes the average rank of each method. The best result is marked in \textbf{bold}.}
\label{tab:main_table}
\begin{tabular}{@{}lccccccc@{}}
\toprule
                     & \multicolumn{1}{l}{PicVocab}  & \multicolumn{1}{l}{ReadEng}     & \multicolumn{1}{l}{PicSeq}  & \multicolumn{1}{l}{ListSort} & \multicolumn{1}{l}{CardSort} & \multicolumn{1}{l}{Flanker} & \multicolumn{1}{l}{Average Rank}  \\ \midrule
\texttt{GCN} (Original Graphs)             & 52.7\tiny\(\pm\)3.77         & 55.4\tiny\(\pm\)3.51            & 51.9\tiny\(\pm\)2.18         & 52.1\tiny\(\pm\)2.55         & 56.6\tiny\(\pm\)6.50         & 48.91\tiny\(\pm\)5.83    & -\\
\midrule
% [0.5ex]
% \cdashline{2-9} 
% \noalign{\vskip 0.5ex}
\texttt{Grad-Indi}            & 53.4\tiny\(\pm\)1.65         & 53.7\tiny\(\pm\)9.48            & 49.3\tiny\(\pm\)3.71         & 48.7\tiny\(\pm\)6.94         & 46.9\tiny\(\pm\)4.65         & 50.7\tiny\(\pm\)2.76     & 7.67\\
\texttt{Grad-Joint}          & 57.8\tiny\(\pm\)3.34         & 58.2\tiny\(\pm\)3.08            & 50.1\tiny\(\pm\)6.17         & 48.9\tiny\(\pm\)5.10         & 52.4\tiny\(\pm\)5.02         & 51.5\tiny\(\pm\)3.94     & 4.33\\
\texttt{Grad-Trained}         & 55.5\tiny\(\pm\)5.29         & 60.0\tiny\(\pm\)1.36            & 49.5\tiny\(\pm\)4.12         & 50.2\tiny\(\pm\)2.20         & 56.3\tiny\(\pm\)7.66         & 51.6\tiny\(\pm\)4.03     & 3.83\\
\texttt{GNNExplainer-Indi}    & 49.7\tiny\(\pm\)3.86         & 55.3\tiny\(\pm\)4.06            & 48.9\tiny\(\pm\)3.29         & 44.8\tiny\(\pm\)3.76         & 52.1\tiny\(\pm\)3.86         & 47.3\tiny\(\pm\)1.58   &  8.33 \\
\texttt{GNNExplainer-Joint}  & 56.4\tiny\(\pm\)7.94         & 55.8\tiny\(\pm\)7.33            & 52.0\tiny\(\pm\)2.84         & 50.1\tiny\(\pm\)3.01         & 53.5\tiny\(\pm\)8.32         & 50.3\tiny\(\pm\)5.81     & 4.67\\
\texttt{GNNExplainer-Trained} & 56.8\tiny\(\pm\)3.10         & 59.2\tiny\(\pm\)2.96            & 51.4\tiny\(\pm\)3.51         & 51.2\tiny\(\pm\)2.01         & 56.0\tiny\(\pm\)4.71         & 50.9\tiny\(\pm\)2.01      & 3.17\\

\texttt{BrainNNExplainer} &57.0\tiny\(\pm\)3.77  &55.7\tiny\(\pm\)5.76 &50.3\tiny\(\pm\)1.47 &49.8\tiny\(\pm\)4.47 &52.4\tiny\(\pm\)3.63 &50.9\tiny\(\pm\)3.95 &4.83\\   

\texttt{BrainGNN} &53.0\tiny\(\pm\)3.25 &47.5\tiny\(\pm\)3.00 &50.7\tiny\(\pm\)3.13 &50.9\tiny\(\pm\)3.13 &50.1\tiny\(\pm\)1.12 &49.0\tiny\(\pm\)6.22 &6.67 \\ 

% IGS (asym)           & \textbf{58.2}\tiny\(\pm\)1.94         & 57.8\tiny\(\pm\)6.45            & 49.0\tiny\(\pm\)3.35         & 51.7\tiny\(\pm\)5.89         & 55.3\tiny\(\pm\)5.89         & 48.3\tiny\(\pm\)1.43     &  4.3\\
\texttt{IGS}            & \textbf{57.8}\tiny\(\pm\)3.10         & \textbf{60.1}\tiny\(\pm\)2.78            & \textbf{53.0}\tiny\(\pm\)4.66         &  \textbf{51.8}\tiny\(\pm\)2.12         & \textbf{57.0}\tiny\(\pm\)5.49         & \textbf{52.1}\tiny\(\pm\)1.97  & 1.00\\ \bottomrule

% Final Sparsity       & 20.4                         & 32.5                            & 35                           & 30                           & 22.5                         & 25.8   \\ \bottomrule
\end{tabular}
\end{table*}

\subsubsection{Baselines} 
% \textcolor{purple}{do you think it to be better to underline or italic?}

\label{subsubsec:baselines}
\revision{We outline the baselines used in our experiments.}
% In this subsection, we explain what our baselines are and how we implement them in our iterative sparsification framework. Before diving into the details of each method, we specify the common notations first: at iteration \(i\), the input graph is denoted as \(\mathcal{G}_i\). The training set is denoted as \(\mathcal{D}_{\text{train}}\), the GNN model is denoted as \(f\), and the importance mask is denoted as \(\mathcal{M}\). We explain what takes place in each iteration below:
% % \xiang{Please add citations to each of your baselines, at least the GNNExplainer and Saliency map. }

\paragraph{Grad-Indi~\citep{sa_individual}} 
\yy{This method obtains the edge importance mask for \ul{each individual graph} from a \ul{trained} GNN model. In contrast to the gradient information \revision{(Strategy S3)} proposed in \Cref{subsec: global_grad}, a gradient map of each sample is generated for the predicted class $C_i$: $\mathbf{T}_i=\nabla f_{C_i}(G_i) \odot \nabla f_{C_i}(G_i)$~\citep{sa_individual}. Later, the edge importance mask $\mathcal{M}_{i}$ for $G_i$ is generated based on \Cref{eq:threshold}.}
% This approach is done post-training and generates masks with respect to each individual sample. At iteration \(i\), first, the GNN model \(f\) is trained until convergence using the input graph \(\mathcal{G}_i\). Next, importance mask \(\mathcal{M}\) is generated for the prediction of a differentiable function $f$ using the squared value of the gradients with respect to input $\mathcal{G}_{i}$ for each individual sample~\citep{sa_individual}: \( \mathcal{M}_j = \lVert \nabla f(\mathcal{G}_{ij})\rVert^2, j \in \{\text{whole dataset index set}\} \). Note that only the gradient with respect to the argmax of model prediction will be kept. At each iteration, each input graph \(\mathcal{G}_{ij}\) is pruned according to thresholding its corresponding importance mask \( \mathcal{M}_{j}\). 

\paragraph{Grad-Joint} 

\revision{We adapt Grad-Indi~\citep{sa_individual} to incorporate our proposed strategies (S1+S3)} and learn an edge importance mask \emph{shared by all graphs} from a \emph{trained} GNN model. Specifically, we leverage the method described in \Cref{subsec: global_grad} that generates the joint gradient map to obtain the joint importance mask.

\paragraph{Grad-Trained}

\revision{We further modify Grad-Indi~\citep{sa_individual} to train the joint edge mask concurrently with the GNN training (S2). We also use the joint gradient map (\Cref{subsec: global_grad}) to initialize the edge importance mask (Strategies S1+S2+S3). The main differences of \texttt{Grad-Trained} from \texttt{IGS} are that: (1) it does not require symmetry of the edge mask; (2) it does not require edge mask sparsity (without $l_1$ regularization).}

% \yy{This method \ul{trains the joint edge mask simultaneously with the training of a GNN and is a variant of \texttt{IGS}.} It uses the joint gradient map (\Cref{subsec: global_grad}) to initialize the edge importance mask. The main differences of this approach from \texttt{IGS} are that: (1) it does not require symmetry of the edge mask; (2) it does not require sparsity of the edge mask (without $l_1$ regularization)}

% This approach is done during-training and generates a joint mask. At each iteration, a trainable joint mask of the size same as each graph input will be initialized and trained together with the GNN model. During training, a strict thresholding of the edge mask \(\mathcal{M}\) is applied to set a certain smallest amount of the edge mask to be zero. Note that this threshold is set to be the same value as our desired pruning threshold. Next, before feeding the input graph into the GNN model, an element-wise product is performed between the trainable edge mask \(\mathcal{M}\) and the input graph \(\mathcal{G}_i\) to generate the sparsified graph and the subsequent training follows as a standard empirical risk minimization process. After the GNN model is trained until convergence, an importance mask $\mathcal{M'}$ is generated using the same procedures as explained in \emph{Gradient-joint}. This $\mathcal{M'}$ is then used to initialize the joint mask in the next iteration. 

\paragraph{GNNExplainer-Indi~\citep{ying2019gnnexplainer}} \yy{This method trains an edge important mask for \ul{each individual graph after the GNN model is trained}. We follow the code provided by \cite{liu2021dig}.}
% This approach is done post-training and generates masks with respect to each individual sample. At iteration \(i\), first, the GNN model \(f\) is trained until convergence using the input graph \(\mathcal{G}_i\). Next, importance mask \(\mathcal{M}\) is generated for the prediction of a differentiable function $f$ using the squared value of the gradients with respect to input $\mathcal{G}_{i}$ for each individual sample using the algorithm from GNNExplainer. At each iteration, each input graph \(\mathcal{G}_{ij}\) is pruned according to thresholding its corresponding importance mask \( \mathcal{M}_{j}\).

\paragraph{GNNExplainer-Joint} 
\yy{Adapted from~\cite {ying2019gnnexplainer}, this model trains \ul{a joint edge important mask for all graphs } \revision{(Strategy S1)}.}
% This approach is done post-training and generates a joint mask. After training a converged GNN model, this time we only train one mask for all the training samples as opposed to the instance-level GNNExplainer that trains different masks for different samples. We have the same objective function as the original algorithm of GNNExplainer to be minimized. The trained mask is then used to prune the whole dataset.

\paragraph{GNNExplainer-Trained} 
\yy{Adapted from~\cite {ying2019gnnexplainer}, this method simultaneously trains \ul{a joint edge important mask and the GNN model} \revision{(Strategies S1+S2)}.} \revision{Compared with \texttt{IGS}, this method does not use gradient information.}

\paragraph{BrainNNExplainer~\cite{cui2021brainnnexplainer}} This method (also known as IBGNN) trains a \ul{joint edge important mask for all graphs after the GNN is trained}. It is slightly different from \texttt{GNNExplainer-Joint} in terms of objective functions. We follow the original setup in \cite{cui2021brainnnexplainer}.

\paragraph{BrainGNN~\cite{braingnn}} This method does not explicitly perform the graph sparsification task, but \ul{uses node pooling to identify important subgraphs}. It learns to preserve important nodes and all the connections between them. We follow the original setup in \cite{braingnn}.

% This approach is done post-training and generates a joint mask. Similar to \emph{Gradient-joint}, a trainable mask will be initialized and trained together with the GNN model. Nevertheless, at each iteration the mask is initialized using the xaiver normal initializer. During training, a  strict thresholding of the edge mask \(\mathcal{M}\) is applied to set a certain smallest amount of the edge mask to be zero. Note that this threshold is set to be the same value as our desired pruning threshold. An elementwise product is performed between the input graph and the trainable edge mask before feeding the graph into the GNN model. We remove the formulation of mutual information in the objective function since this is done during training while keeping every other detail the same. Once trained until convergence, the trained mask can be used to prune the input graph for the next iteration. 

% \reminder{Explain what are the different baselines. for example, saliency version, sliency training version, what you modify, what is the main difference. similar for the other baselines, give the details, especially where do the variants differ. }

\subsubsection{Training Setup.} 
% We manually construct the training setup from the original dataset provided by ConnectomeDB. For each of the cognitive labels, we formulate the corresponding binary classification task as tail/head prediction. We treat the samples with scores in the top 1/3 percentile range as one class and the samples with scores in the lowest 1/3 percentile range as the other class. Together they construct the dataset for a certain prediction task. 
\yy{To fairly evaluate different methods under the iterative framework, we adopt the same GNN architecture ~\citep{gcn}, hyper-parameter settings, and training framework. 
We set the number of convolutional layers to four, the dimension of the hidden layers to 256, the dropout rate to 0.5, the batch size to \revision{16}, the optimizer to Adam, the learning rate to 0.001, and the regularization coefficient $\lambda$ to \revision{0.0001}. }Note that though we use the GNN from~\citep{gcn}, \texttt{IGS} is model-agnostic, and we provide the results of other backbone GNNs in \Cref{tab:different_model_IGS}.
For each prediction task, we shuffle the data and take four different data splits. The train/val/test split is 0.7/0.15/0.15. To reduce the influence of imbalances, we manually ensure each split has equal labels. 
% We report the test accuracy averaged over the four splits.
% that may occur by the random shuffling from the class with more samples 
% in each split. 
% For a fair comparison, we use the exactly same GNN model throughout different baseline/method runs. We use the graph convolutional network~\citep{gcn} following the implementation from PyTorch-Geometric~\citep{torch_geometric} as our GNN model. Throughout the settings, we fix the batch size to 16, dropout to 0.5, learning rate to 1e-3, and the number of convolutional layers to 4 with the same hidden channel dimensions of 256. 
\yy{In each iteration, we
adopt early stopping~\citep{prechelt2012early} and set the patience to 100 epochs. We stop training if we cannot observe a decrease in validation loss in the latest 100 epochs.}
% set a patience epoch range of 100 and we use the smallest validation loss as the stopping criterion. 
We fix \yy{the removing ratio $p\%$} to be 5\% per iteration. In the iterative sparsification, we run a total of 55 iterations and use the validation loss of the sparsified graphs as the criterion to select the best iteration (Step 3 in Figure ~\ref{fig:trained_mask}). 
We present the average and standard deviation of test accuracies over four splits, using the model obtained from the best iteration. \textbf{The code is available at \url{https://github.com/motivationss/IGS.git}. }

% \reminder{how do you maintain the balance of class labels, how many splits, train/val/test percentage. How do you average the results, what metric you use to evaluate(accuracy, std), how do you decide which iteration to stop, do you set the same parameters for all the baselines? learning rate and other necessary details to reproduce your results.}
\subsection{(Q1-Q3) Graph Classification under the Iterative Framework}
\label{subsec: iterative_framework}
In \Cref{tab:main_table}, we present the results of \texttt{IGS} with the eight baselines mentioned in \cref{subsubsec:baselines}. The first row represents the prediction task we study; the second row represents the performance averaged across four different splits using the original graph; and the rest of the rows denote the performance of other baselines. Notably, for better comparison across different baselines, the last column shows the average rank of each method. \revision{Below we present our observations }from \Cref{tab:main_table}:

First, learning a joint mask contributes to a better performance than learning a mask for each graph separately. We can start by comparing the performance between \texttt{GNNExplainer-Joint} and \texttt{GNNExplainer-Indi} as well as \texttt{Grad-Joint} and \texttt{Grad-Indi}. The performance disparity between the methods in each pair is notable and consistent across all prediction tasks. Notably, \texttt{Grad-Joint} \revision{(rank: 4.33)} outperforms \texttt{Grad-Indi} \revision{(rank: 7.67)} by a considerable margin, while \texttt{GNNExplainer-Joint} (rank: 4.67) ranks significantly higher than \texttt{GNNExplainer-Indi} \revision{(rank: 8.33)}. Using a joint mask instead of individual masks can provide up to $6.7 \%$ boost in accuracy, validating our intuition in \cref{subsec: global_grad} that a joint mask is more robust to sample-wise noise. 

Second, training the mask and the GNN model simultaneously yields better results than obtaining the mask from the trained model. We can see this by comparing the performance between \revision{the \texttt{Trained} and the \texttt{Joint} variants of \texttt{Grad} and \texttt{GNNExplainer}}. Changing from post-training to joint-training can provide up to $3.4 \%$ performance improvements, as demonstrated in the ReadEng task by the two variants of \texttt{GNNExplainer}. Even though in some tasks the post-training approach may outperform the trained approach (\emph{e.g.} \texttt{Grad-Joint} and \texttt{Grad-Trained} in the PicVocab task), the trained approach has a higher average rank than the post-training approach (\emph{e.g.} 3.83 vs. 4.33 \revision{for \texttt{Grad}} and 3.17 vs. 4.67 \revision{for \texttt{GNNExplainer}}). In addition, the better performance of \texttt{IGS} over \texttt{BrainNNExplainer} also demonstrates the effectiveness of obtaining the edge mask during training rather than after training.

{Third, incorporating gradient information helps improve classification performance. We can see this by first comparing the performance of \texttt{Grad-Joint} and \texttt{Grad-Trained} against the original graphs. The use of gradient information can provide up to 5.1\% \revision{higher accuracy}, though the improvement \revision{depends on the task}. Furthermore, since the main difference between \texttt{GNNExplainer-Trained} and \texttt{IGS} lies in the use of gradient information, the consistent superior performance of \texttt{IGS} strengthens this conclusion.}

Fourth, we compare the performance \revision{of the baselines} against the performance of the original graphs (second row). \texttt{Grad-Indi}~\citep{sa_individual} and \texttt{GNNExplainer-Indi}~\cite{ying2019gnnexplainer} are implementations that faithfully follow their original formulation or are provided directly by the authors. 
% directly follows from the published Github implementation, 
These two approaches fail to achieve any performance improvements through iterative sparsification, with the exception of \texttt{Grad-Indi} in the task of PicVocab and ReadEng. This raises the question of whether these existing instance-level approaches can identify the most meaningful edges in noisy graphs. These methods may be vulnerable to severe sample-wise noise. On the contrary, with our suggested modifications, the joint and trained versions can remove the noise and provide up to $5.1\%$ performance boost compared to the \revision{base GCN method applied to the} original graphs. However, the improvement is \revision{dataset-dependent}. For instance, \texttt{GNNExplainer-Trained} provides decent performance boosts in PicVocab, ReadEng, and Flanker, but degrades in PicSeq, ListSort, and CardSort.

Finally, our proposed approach, \texttt{IGS}, achieves the \textbf{best} performance across all prediction tasks, demonstrated by its highest rank among all methods. Compared with the performance \revision{on} the original graphs, \texttt{IGS} can provide consistent performance boost across all prediction tasks, with the exception of ListSort, which is a challenging task that no baseline surpasses the original performance. Furthermore, using the sparsified graph identified by \texttt{IGS} generally results in less variance in accuracy and leads to better stability when compared to the original graphs, with the exception on the PicSeq task. {In addition, the superior performance of \texttt{IGS} over \texttt{BrainGNN} demonstrates the effectiveness of using edge importance masks as opposed to node pooling.}

\vspace{0.1cm}
\noindent \revision{\textit{Graph Sparsity}.} In \Cref{tab:sparsity}, we present the final average sparsity of the graphs obtained by \texttt{IGS} over four data splits. We observe that with significantly fewer edges retained, \texttt{IGS} can still achieve up to $5.1\%$ performance boost.

\begin{table}[!h]
% \centering
\caption{Final sparsity of the sparsified brain graphs identified by \texttt{IGS} averaged over different splits. The initial sparsity is 50\% \revision{by thresholding}. \texttt{IGS} can remove \revision{more than} half of the edges while achieving up to 5.1\% performance boost.}
\label{tab:sparsity}
\resizebox{0.5\textwidth}{!}{ 
\begin{tabular}{@{}lllllll@{}}
\toprule
    & PicVocab                 & ReadEng                  & PicSeq                   & ListSort               & CardSort                 & Flanker                \\ \midrule
Sparsity(\%) & \multicolumn{1}{c}{22.5} & \multicolumn{1}{c}{35.5} & \multicolumn{1}{c}{35.5} & \multicolumn{1}{c}{30.0} & \multicolumn{1}{c}{25.0} & \multicolumn{1}{c}{25.0} \\ \bottomrule
\end{tabular}
}
\end{table}

\begin{figure*}[!ht]
\minipage{0.33\textwidth}
    \includegraphics[width=\columnwidth]{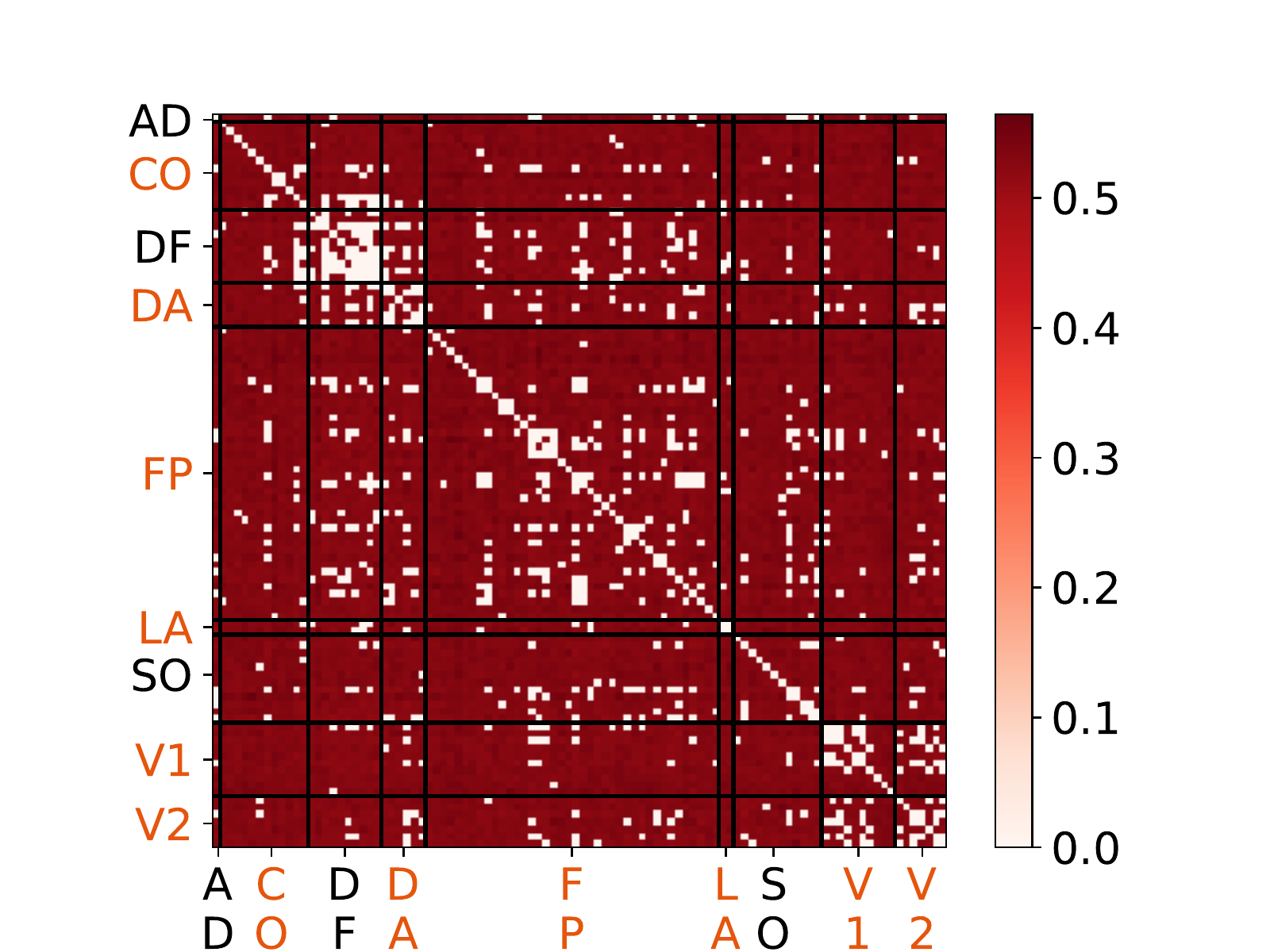}
\endminipage
% \hfill
\minipage{0.33\textwidth}
    \includegraphics[width=\columnwidth]{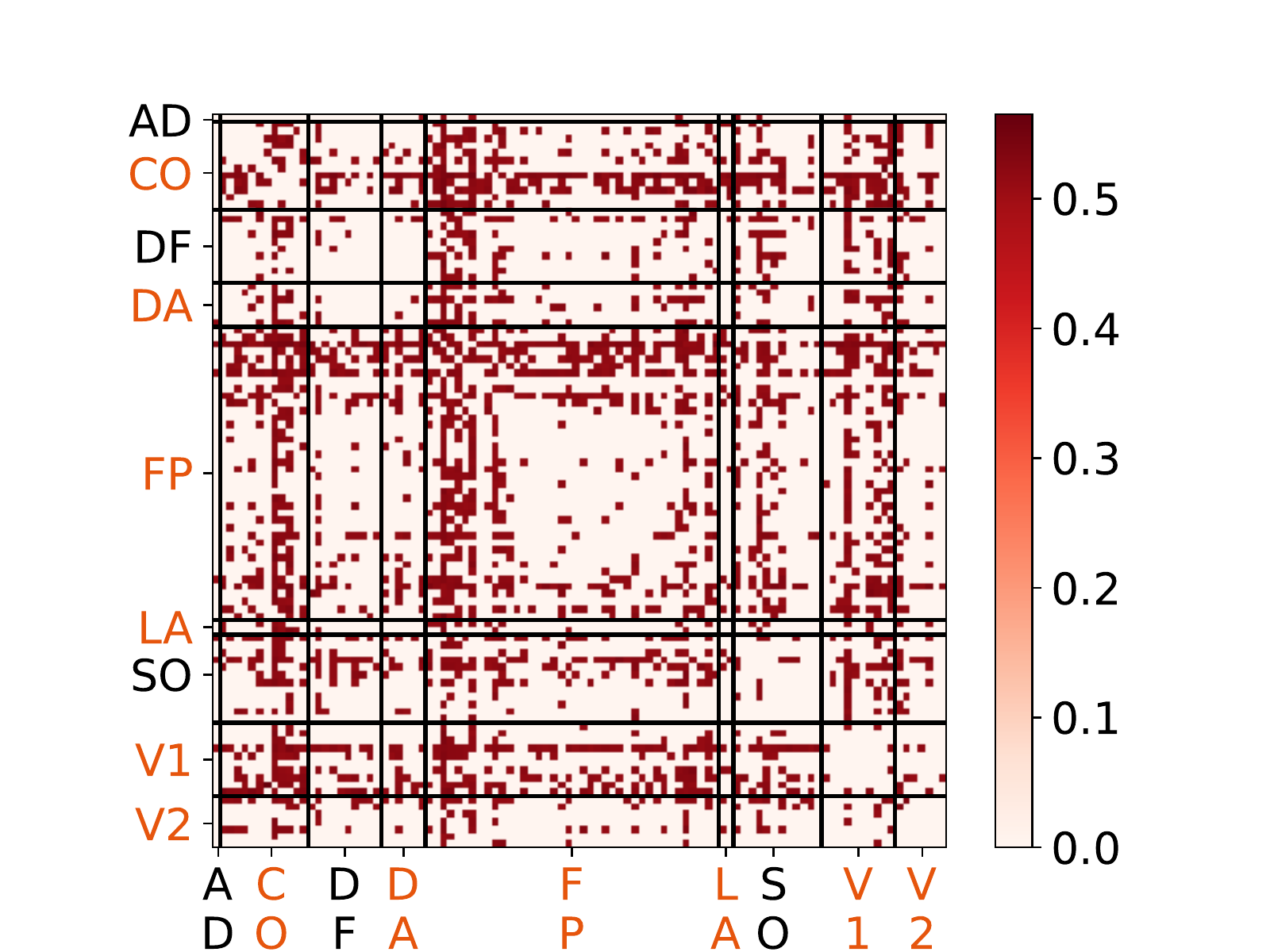}
\endminipage
% \hfill
\minipage{0.33\textwidth}
    \includegraphics[width=\columnwidth]{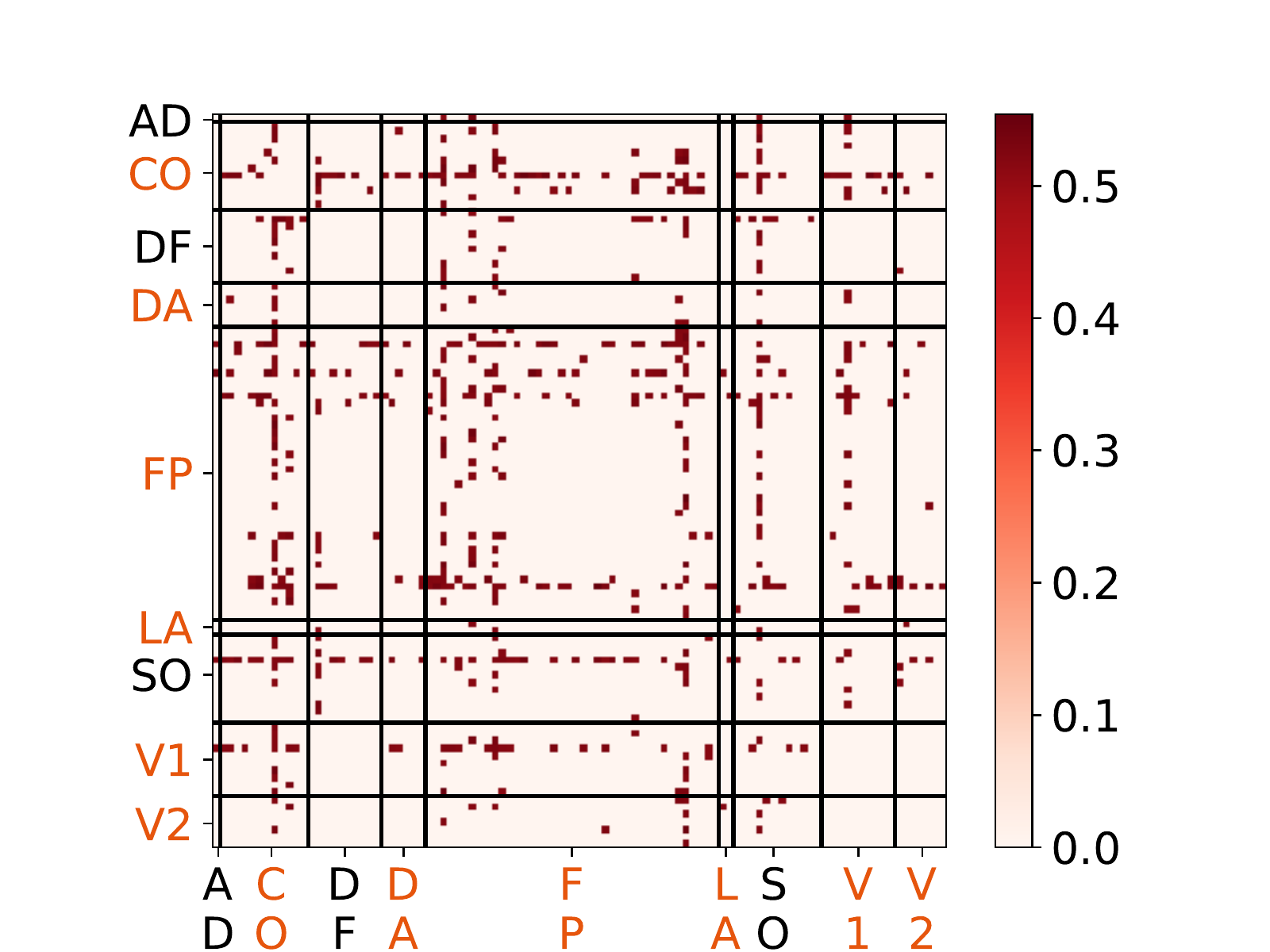}
\endminipage
\\ 
\minipage{0.33\textwidth}
    \includegraphics[width=\columnwidth]{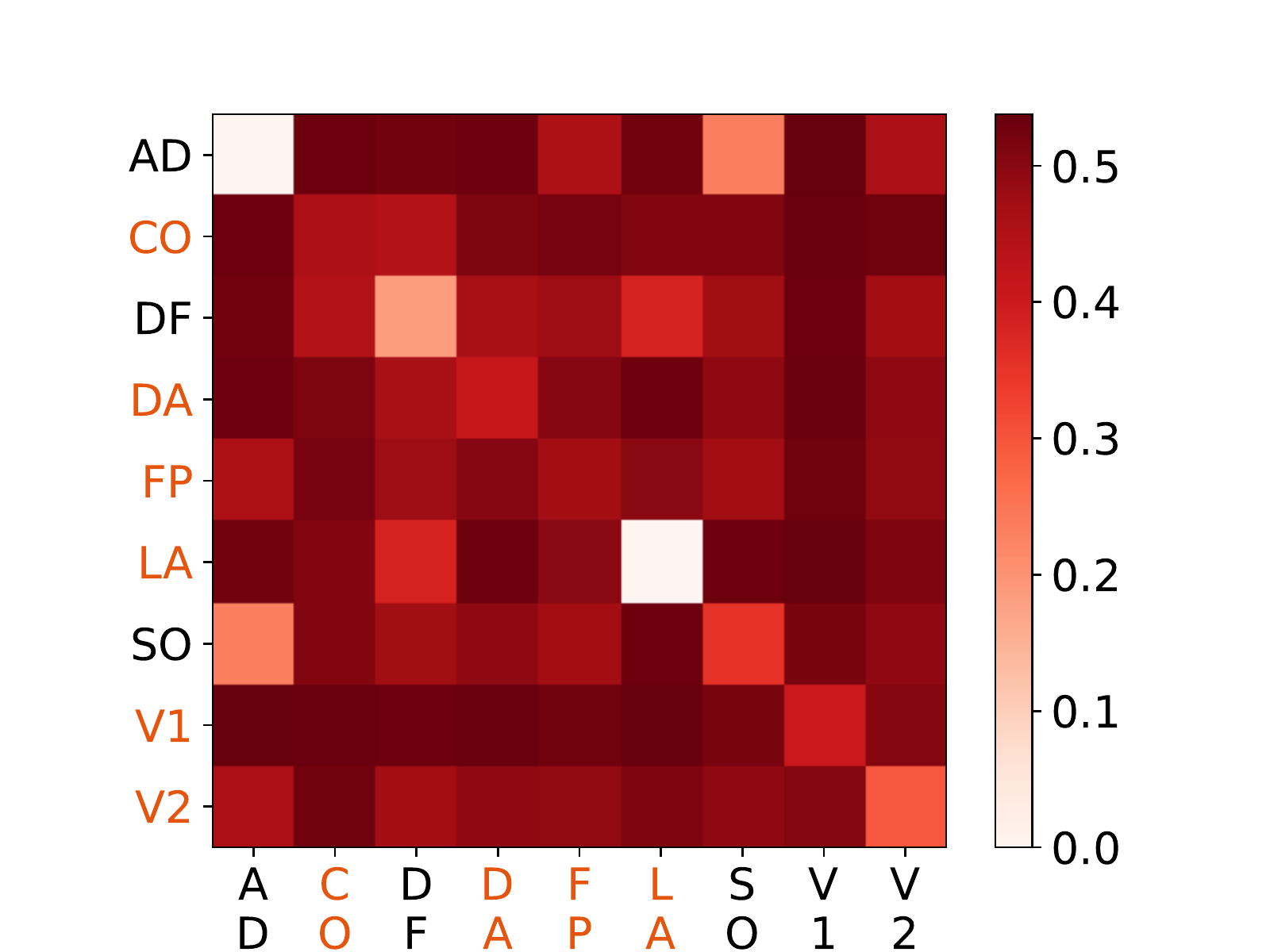}
\endminipage
\minipage{0.33\textwidth}
    \includegraphics[width=\columnwidth]{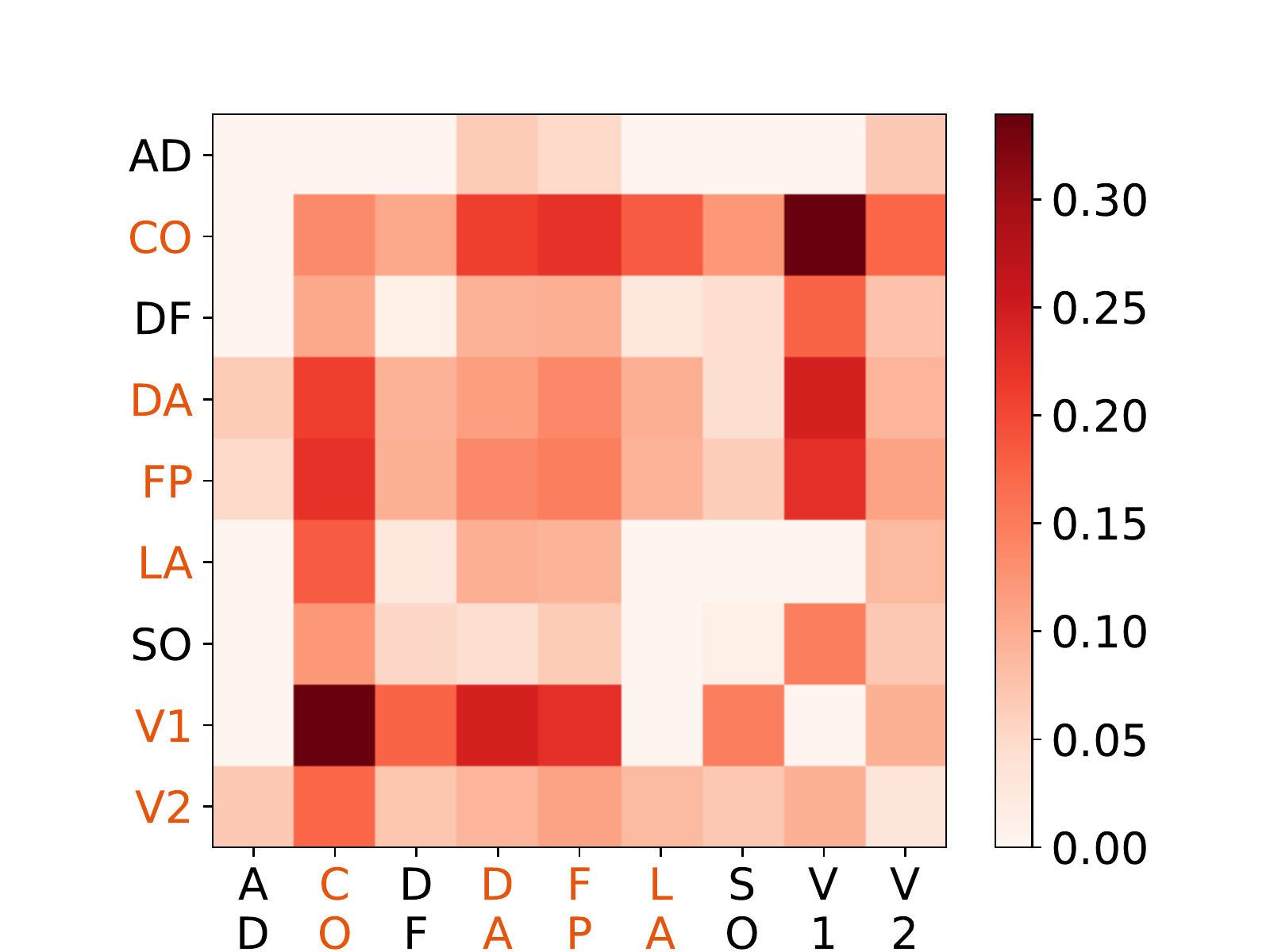}
\endminipage
\minipage{0.33\textwidth}
    \includegraphics[width=\columnwidth]{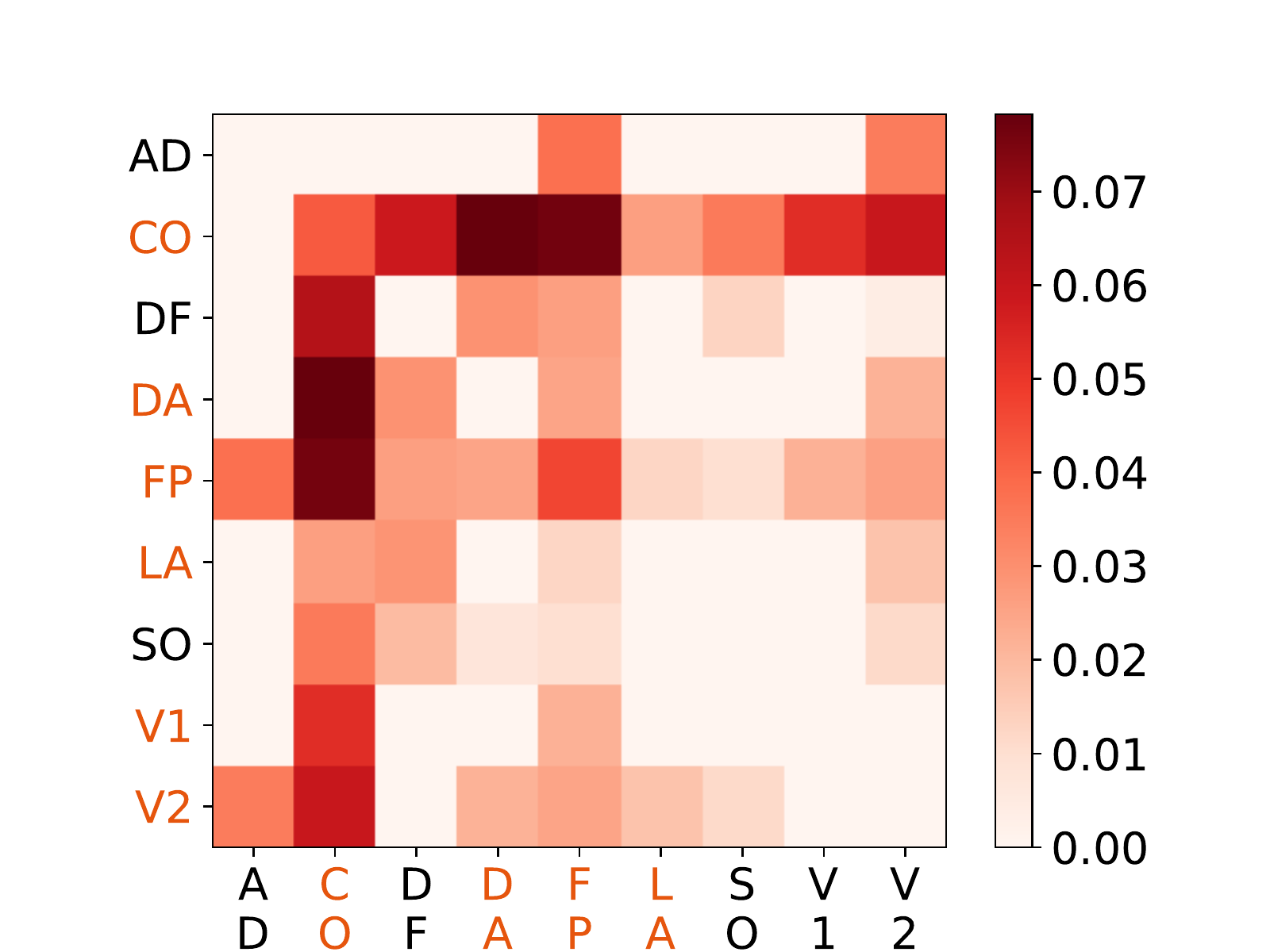}
\endminipage
    \caption{Weighted brain network edge masks at both node (top row) and subnetwork level (bottom row - computed as the average of corresponding edges) for PicVocab task. Early, middle, and final phases of training are depicted from left to right, and high-importance subnetworks are highlighted in red. We find that \texttt{IGS} gradually removes noisy edges and its final edge importance mask can provide high-quality interpretations. Highlighted (Orange) label names represent the regions that are meaningful in this task. Brain network labels and abbreviations:  Auditory (AD), Cingulo-Opercular (CO), Dorsal Attention (DA), Default (DF), Frontoparietal (FP), Language (LA), Somatomotor (SO), Visual 1 (V1), Visual 2 (V2).}
    \label{fig:PicVocab_interpretation}
    % \vspace{-3mm}
\end{figure*}

% \begin{figure*}[!ht]
% \minipage{0.33\textwidth}
%     \includegraphics[width=\columnwidth]{Fig/100ReadEngEarly.pdf}
% \endminipage
% % \hfill
% \minipage{0.33\textwidth}
%     \includegraphics[width=\columnwidth]{Fig/100ReadEngMiddle.pdf}
% \endminipage
% % \hfill
% \minipage{0.33\textwidth}
%     \includegraphics[width=\columnwidth]{Fig/100ReadEngFinal.pdf}
% \endminipage
% \\ 
% \minipage{0.33\textwidth}
%     \includegraphics[width=\columnwidth]{Fig/GeneralReadEngEarly.pdf}
% \endminipage
% \minipage{0.33\textwidth}
%     \includegraphics[width=\columnwidth]{Fig/GeneralReadEngMiddle.pdf}
% \endminipage
% \minipage{0.33\textwidth}
%     \includegraphics[width=\columnwidth]{Fig/GeneralReadEngFinal.pdf}
% \endminipage
%     \caption{Weighted brain network edge masks at both node (top row) and subnetwork level (bottom row) for ReadEng task, following the same convention and having the same conclusions as \Cref{fig:PicVocab_interpretation}.}
%     \label{fig:ReadEng_interpretation}
% \end{figure*}

% Adding Figures right now  --- Oliver ReadEng and PicVocab according to what you write here
% Hey Marlena Do you want the progression of the 100 x 100 matrix or the general one? --- Oliver

\subsection{(Q4) Interpretability of IGS}
\label{exp:interpretability}
We \revision{now} evaluate the interpretability of the edge masks derived for each of our prediction tasks. 

\vspace{0.1cm}
\noindent \revision{\textit{Setup}. We assign} anatomical labels to each of the 100 components comprising the nodes of our brain networks by computing the largest overlap between regions identified in the Cole-Anticevic parcellation~\cite{ji2019mapping}. We then obtained the edge masks from the best-performing iteration of each prediction task and assessed the highest-weighted edges in each mask. 

\vspace{0.1cm}
\noindent 
\revision{\textit{Results.}} Since our \texttt{IGS} model performed best in the language-related prediction tasks, ReadEng and PicVocab, we focus our interpretability analysis on this domain. There is ample evidence in the neuroscience literature that supports the existence of an intrinsic language network that is perceptible during resting state~\cite{tomasi2012resting, klingbeil2019resting, branco2020mapping}; thus, it is unsurprising that our rs-fMRI based brain networks are predictive of language task performance. It has also been well established for over a century that the language centers (including Broca's area, Wernicke's area, the angular gyrus, etc.) are characteristically left-lateralized in the brain~\cite{broca1861remarques, wernicke1874aphasische}. In both ReadEng and PicVocab, the majority of the highest weighted edges retained in the masks involved brain regions localized to the left hemisphere, falling in line with the expectations for a language task.

\revision{\textit{PicVocab.}} \Cref{fig:PicVocab_interpretation,fig:ReadEng_interpretation} depict the progression of the edge masks at both the node and subnetwork level over the training iterations towards optimal edge mask in both the ReadEng and PicVocab tasks. Evaluating the edge masks at the subnetwork level offers valuable insights into which functional connections are most important for the prediction of each task. The PicVocab edge mask homed in on functional connections involving the Cingulo-Opercular (CO) network, specifically between CO and the Dorsal Attention (DA), Visual1 (V1), Visual2 (V2) and Frontoparietal (FP) networks. The CO network has been shown to be implicated in word recognition~\cite{vaden2013cingulo}, and its synchrony with other brain networks identified here may represent the stream of neural processing related to the PicVocab task, in which subjects respond to an auditory stimulus of a word and are prompted to choose the image that best represents the word. Connectivity between the Auditory (AD) and V2 networks is also evident in the PicVocab edge mask, suggesting the upstream integration of auditory and visual stimuli involved in the PicVocab task are also predictive of task performance.

\revision{\textit{ReadEng.}} The \texttt{IGS} model also found edge mask connections between the V1 network and the CO, Language (LA) and DA networks, as well as CO-LA and CO-AD connections, to be most predictive of ReadEng performance. This task involves the subject reading aloud words presented on a screen. From our results, it follows that the ability of Vis1 to integrate with networks responsible for language processing (LA and CO) and attention (DA), as well as the capacity for functional synchrony between the language-related networks (CO-LA), would be predictive of overall ReadEng performance. The importance of the additional CO-AD connectivity identified by our model also suggests that the ability of the CO language network to integrate with auditory centers may be involved in the neural processes responsible for the proper pronunciation of the words given by visual cues. 

\noindent 
\revision{\textit{Key take-aways.}} Overall, in addition to the \texttt{IGS} model's superior classification performance, our results suggest that the iterative pruning of the \texttt{IGS} edge masks during training does indeed retain important and neurologically meaningful edges while removing noisy connections. While it has been shown in the literature that resting-state connectivity can be used to predict task performance~\cite{jones2017resting, mennes2010inter, baldassarre2012individual}, the ability of the \texttt{IGS} model to sparsify the resting state brain graph to clearly task-relevant edges for prediction of task performance further underscores the interpretability of the resultant edge masks.

\section{Related Work}
% \label{sec:related_work}
\vspace{3mm}

\vspace{-0.2cm}
\subsection{Graph Explainability} 
\label{sec:related_work}
Our work is related to explainable GNNs \revision{given that we} identify important edges/subgraphs that account for the model predictions.
% An inspiring line of work is explainable graph neural networks (explainable GNN). 
 % identify important subgraphs which can retain similar performance given a well-trained model.  Even though no current explainable GNNs are proposed to experiment on brain graphs, it is worth mentioning them since this direction is closely related to our method and can offer substantial insights. 
 Some explainable GNNs are ``perturbation-based'', where the goal is to investigate the relation between output and input variations. GNNExplainer~\cite{ying2019gnnexplainer} learns a soft mask for the nodes and edges, which explains the predictions of a well-trained GNN model. SubgraphX \cite{subgraphX}
 explains its predictions by efficiently exploring different subgraphs with a Monte Carlo tree search.
 % uses the Monte Carlo Tree Search (MCTS)\cite{MCTS} algorithm to find separate subgraphs via node pruning and identify the most important subgraph from the leaves of the search tree as the explanation of the prediction. 
 % The other popular category 
 Another \revision{approach for} explainable GNNs is surrogate-based\revision{;} the methods in this category generally construct a simple and interpretable surrogate model to approximate the output of the original model \revision{in} certain neighborhoods~\citep{yuan2022explainability}. For instance, GraphLime \cite{graphlime} considers the N-hop neighboring nodes of the target node and then trains a nonlinear surrogate model to fit the local neighborhood predictions; RelEx~\citep{zhang2021relex} first uses a GNN to fit the BFS-generated datasets and then generates soft masks to explain the predictions; PGM-Explainer~\citep{vu2020pgm} generates local datasets based on the influence of randomly perturbing the node features, shrinks the size of the datasets via the Grow-Shrink algorithm, and employs a Bayesian network to fit the datasets. In general, most of these methods focus on the node classification task and make explanations for a single graph, which is not applicable to our setting. 
Others only apply to simple graphs, which cannot handle signed and weighted brain graphs~\cite{graphlime, subgraphX}. Additionally, most methods generate explanations after a GNN is trained.
 % Most of the methods focus on simple unweighed graphs and cannot be directly applied to weighted graphs. 
 Though some methods achieve decent results in explainability-related metrics (\emph{e.g.} fidelity scores~\cite{pope2019explainability}), it remains unclear whether their explanations can necessarily remove noise and retain the ``important'' part of the original graph, which improves the classification accuracy.  
 % Additionally, these methods are all post-training and fail to consider the potential of generating explanations during training.  

% \vspace{3mm}
% \noindent \textbf
\subsection{Graph Sparsification} %\\
Compared to the explainable GNN methods, graph sparsification methods \revision{explicitly aim to sparsify graphs}. Most of the existing methods are unsupervised~\cite{neuralSparse}. Conventional methods reduce the size of \revision{the} graph through approximating pairwise distances~\citep{peleg1989graph}, preserving various kinds of graph cuts~\citep{karger1994random}, node degree distributions~\citep{eden2018provable, voudigari2016rank}, and using some graph-spectrum based approachse~\citep{calandriello2018improved, chakeri2016spectral, adhikari2017propagation}. These methods aim at preserving the structural information of the original input graph without using the \yy{label} information\revision{, and they} assume that the input graph is unweighted. Relatively fewer supervised works have been proposed. \revision{For example,} NeuralSparse~\citep{neuralSparse} builds a parametrized network to learn a k-neighbor subgraph by limiting each node to have at most $k$ edges. On top of NeuralSparse, PTDNet~\citep{PTDNet} removes the k-neighbor assumption, and instead, it employs a low-rank constraint on the learned subgraph to discourage edges connecting multiple communities. Graph Condensation~\citep{jin2021graph} proposes to parameterize the condensed graph structure as a function of condensed node features and optimizes a gradient-matching training objective. Despite the new insights offered by these methods, most of them focus exclusively on node classification, and their training objectives are built on top of that.  {\revision{A} work that shares similarity to our proposed \revision{method}, \texttt{IGS}, is BrainNNExplainer~\citep{cui2021brainnnexplainer} (also known as IBGNN). It is inspired by GNNExplainer~\citep{ying2019gnnexplainer} and obtains the joint edge mask in a post-training fashion. \revision{On the other hand,} our proposed method, \texttt{IGS}, trains a joint edge mask along with the backbone model and incorporates gradient information in an iterative manner. Another line of work \revision{leverages} node pooling to identify important subgraphs\revision{, and} learns to preserve important nodes and all the connections between them. One representative work is BrainGNN \cite{braingnn}. However, the connections between preserved nodes are not necessarily all informative, and some may contain noise.} 

% \vspace{3mm}
% \noindent \textbf
\subsection{Saliency Maps}
% \noindent
Saliency maps are first proposed to explain the deep convolutional neural network models in image classification tasks~\citep{simonyan2013deep}. Specifically, the method proposes to use the gradients backpropagated from the predicted class as the explanations. Recently, \cite{sa_individual} introduces the concept of saliency maps to graph neural networks, employing squared gradients to explain the \revision{underlying} model. Additionally, \cite{arslan2018graph} suggests using graph saliency to identify regions of interest (ROIs).
In general, the gradients backpropagated from the output logits can serve as the importance indicators
% important proxy 
for model predictions. In this work, inspired by the line of saliency-related works, we leverage the gradient information to guide our model.

\section{Conclusions}
In this paper, we stud\revision{ied} neural-network-based graph sparsification for brain graphs. By \revision{introducing} an iterative sparsification framework, we identif\revision{ied} several effective \revision{strategies} for GNNs to filter out noisy edges and improve the graph classification performance. We combine\revision{d} these \revision{strategies} into a new \revision{interpretable graph classification} model, \texttt{IGS}, which improves the graph classification performance by up to 5.1\% with 55\% fewer edges than the original graphs. The retained edges identified by \texttt{IGS} provide neuroscientific interpretations and are
supported by well-established literature.

\section*{Acknowledgements}
{\small We thank the anonymous reviewers for their constructive feedback. This material is based upon work supported by the National Science Foundation under IIS 2212143, CAREER Grant No.~IIS 1845491, a Precision Health Investigator Award at the University of Michigan, 
and AWS Cloud Credits for Research. 
Data were provided [in part] by the Human Connectome Project, WU-Minn Consortium (PIs: D. Van Essen and K. Ugurbil; 1U54MH091657) funded by the 16 NIH Institutes and Centers that support the NIH Blueprint for Neuroscience Research; and by the McDonnell Center for Systems Neuroscience at Washington University. 
Any opinions, findings, and conclusions or recommendations expressed in this material are those of the authors and do not necessarily reflect the views of the National Science Foundation or other funding parties.
}

\nocite{Yang2022SparseAC}
\nocite{Wu_2023_CVPR}

% \newpage

\bibliographystyle{ACM-Reference-Format}
\balance
\bibliography{paper}
\appendix
% \newpage

% \section{Additional Experiment Results}

% \newpage 
\section{IGS with Other Backbone GNNs}

In \Cref{tab:different_model_IGS}, we present the results of \texttt{IGS} evaluated on different GNN backbones (noted by ``- IGS'') and compare it against the original performance (noted by ``Original Graphs''). Specifically, we consider three additional GNN models: GraphSAGE~\citep{graphsage}, GraphConv~\citep{graphconv}, and GIN~\citep{gin}. The experimental and hyperparameter  settings follow those in \Cref{sec:exp_setup}. Compared with the performance of the original graphs, the sparsified graphs obtained from \texttt{IGS} consistently contribute to performance gains across all GNN backbones and prediction tasks. It provides an average of 4.72\% increase in the test accuracies for GraphSage, an average of 1.92\% increase in the test accuracies for GraphConv, and an average of 1.45\% increase in the test accuracies for GIN. This demonstrates that the improvements achieved by \texttt{IGS} are model-agnostic.

% Added

% \usepackage{booktabs}
\begin{table*}[t]
\caption{Performance of \texttt{IGS} with different GNN backbones, following the same setup in \Cref{sec:exp_setup}. The performance improvements achieved by \texttt{IGS} are model-agnostic.}
\label{tab:different_model_IGS}
% \resizebox{0.8\textwidth}{!}{ 
\begin{tabular}{@{}lcccccc@{}}
\toprule
                               & PicVocab                            & ReadEng                             & PicSeq                              & ListSort                            & CardSort                            & Flanker                             \\ \midrule
GraphSage (Original Graphs) & 56.2\tiny\(\pm\)5.47 & 49.6\tiny\(\pm\)2.37 & 48.1\tiny\(\pm\)4.92 & 50.6\tiny\(\pm\)0.63 & 50.3\tiny\(\pm\)1.75 & 49.0\tiny\(\pm\)2.04 \\
GraphSage - IGS                & 60.9\tiny\(\pm\)4.27 & 56.4\tiny\(\pm\)2.27 & 55.1\tiny\(\pm\)3.52 & 52.6\tiny\(\pm\)1.66 & 54.4\tiny\(\pm\)11.8 & 52.7\tiny\(\pm\)2.40 \\ \midrule
GraphConv (Original Graphs) & 53.2\tiny\(\pm\)6.70 & 54.9\tiny\(\pm\)5.06 & 48.2\tiny\(\pm\)1.19 & 49.4\tiny\(\pm\)0.63 & 50.6\tiny\(\pm\)2.93 & 49.4\tiny\(\pm\)2.54 \\
GraphConv - IGS                & 57.1\tiny\(\pm\)8.21 & 55.9\tiny\(\pm\)3.41 & 52.3\tiny\(\pm\)1.93 & 50.7\tiny\(\pm\)2.19 & 50.8\tiny\(\pm\)7.70 & 50.4\tiny\(\pm\)9.37 \\ \midrule
GIN (Original Graphs)       & 55.8\tiny\(\pm\)5.42 & 56.4\tiny\(\pm\)6.94 & 49.9\tiny\(\pm\)3.53 & 52.6\tiny\(\pm\)2.84 & 55.0\tiny\(\pm\)3.00 & 48.5\tiny\(\pm\)3.83 \\
GIN - IGS                      & 59.3\tiny\(\pm\)5.83 & 56.7\tiny\(\pm\)5.54 & 51.0\tiny\(\pm\)3.28 & 54.1\tiny\(\pm\)5.70 & 55.0\tiny\(\pm\)6.47 & 50.8\tiny\(\pm\)4.80 \\ 
\bottomrule
\end{tabular}
% }
\end{table*}

\section{Additional Studies on interpretability}

\begin{figure*}[!ht]
\minipage{0.33\textwidth}
    \includegraphics[width=\columnwidth]{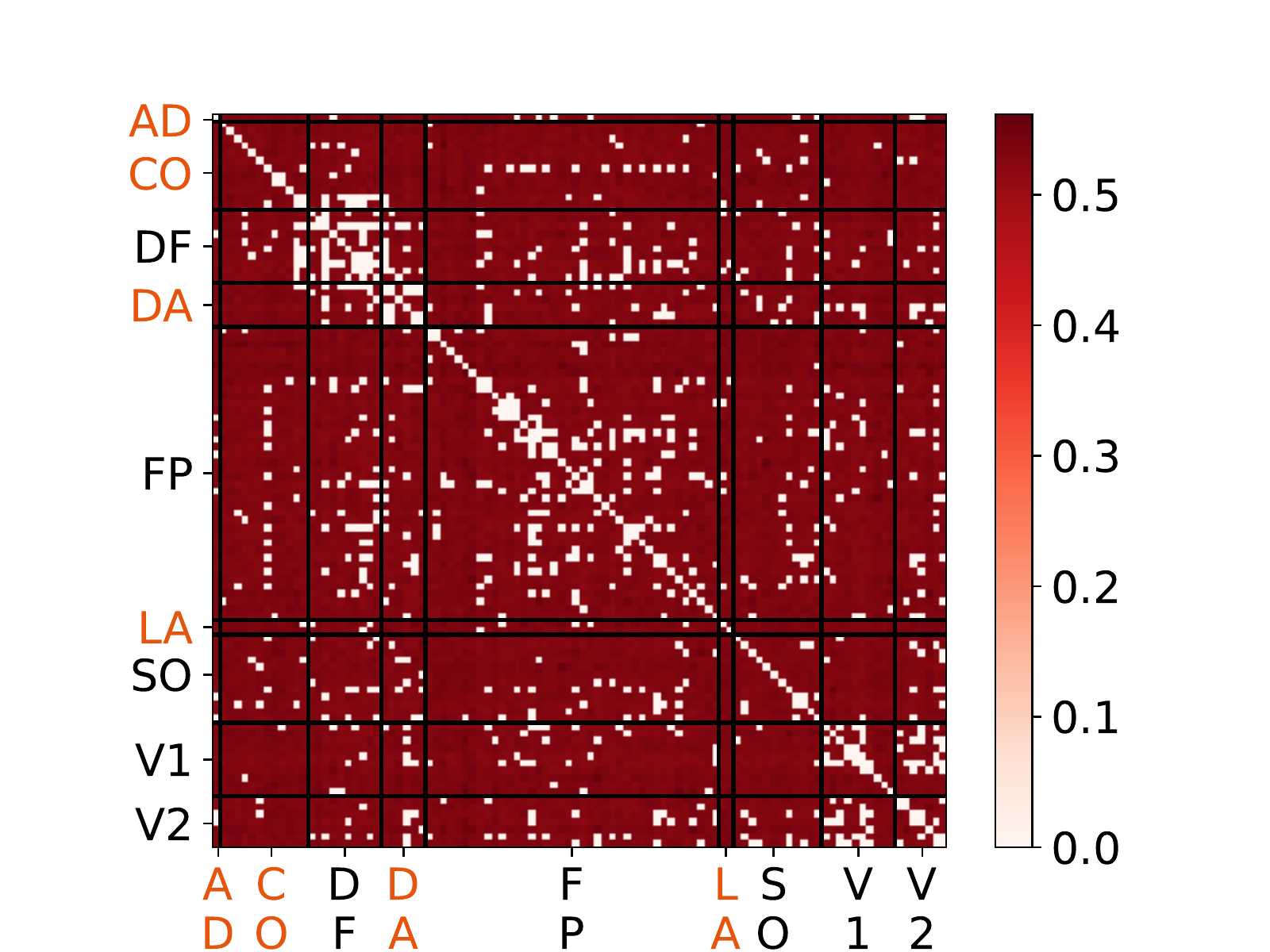}
\endminipage
% \hfill
\minipage{0.33\textwidth}
    \includegraphics[width=\columnwidth]{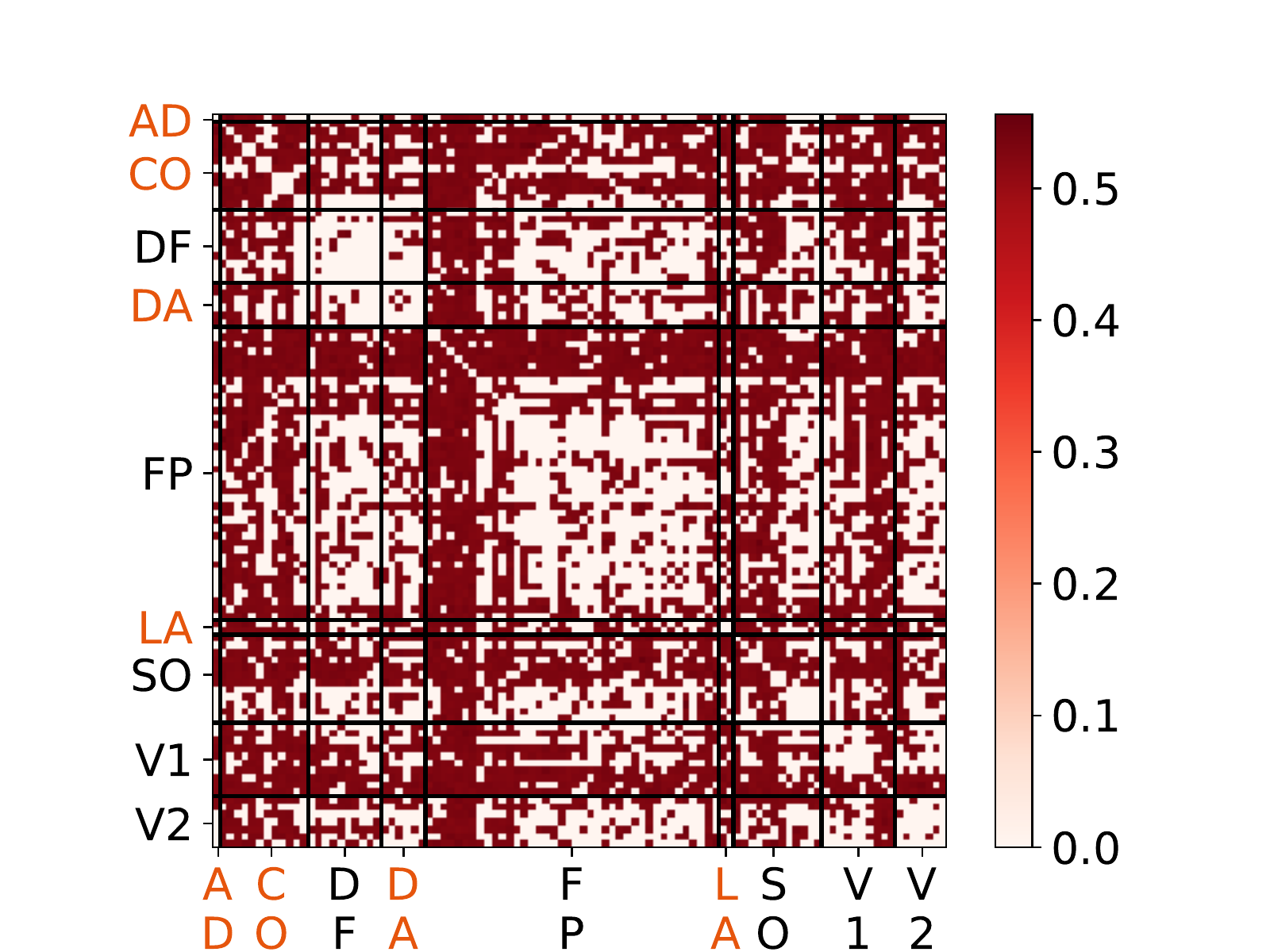}
\endminipage
% \hfill
\minipage{0.33\textwidth}
    \includegraphics[width=\columnwidth]{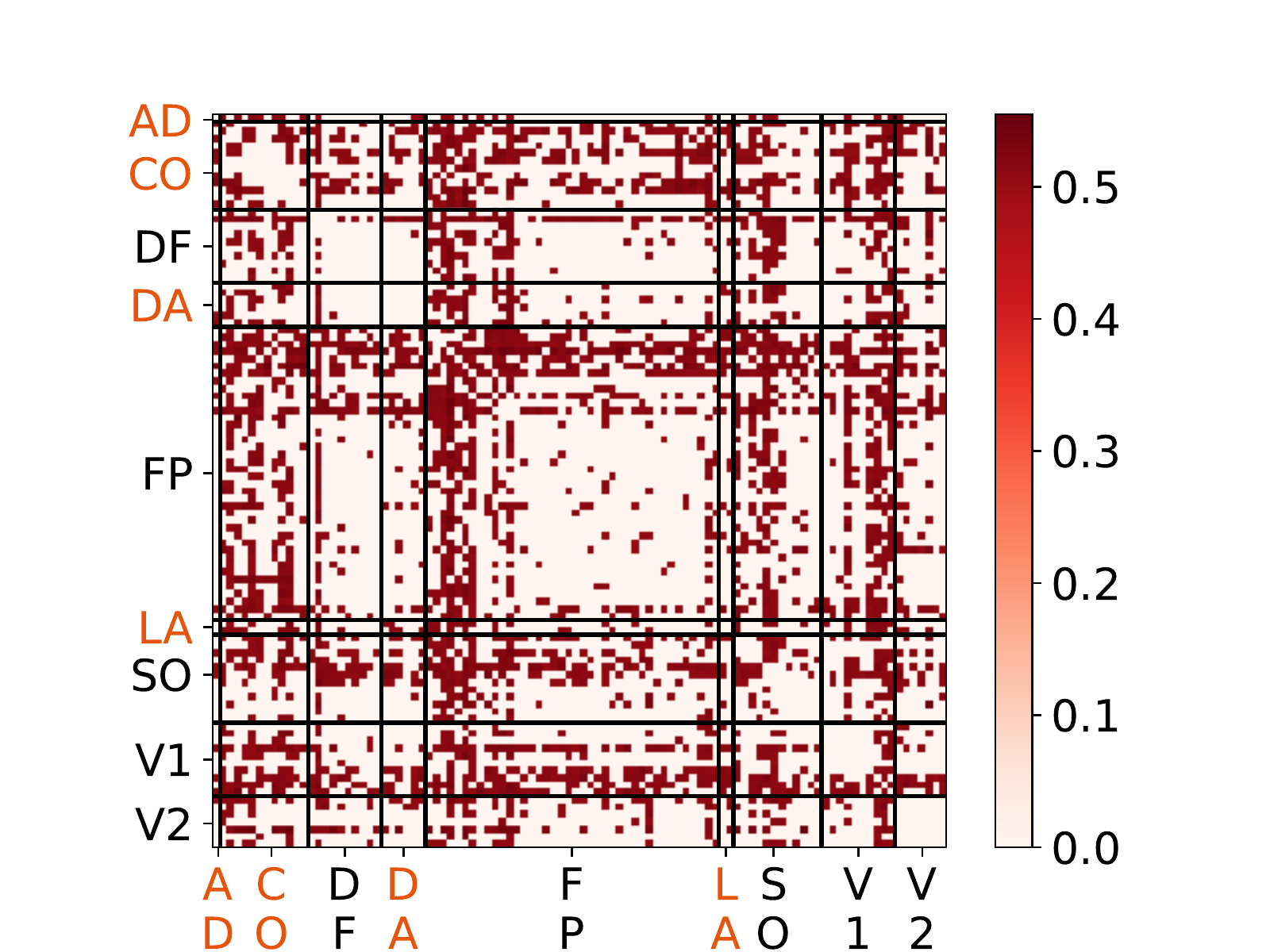}
\endminipage
\\ 
\minipage{0.33\textwidth}
    \includegraphics[width=\columnwidth]{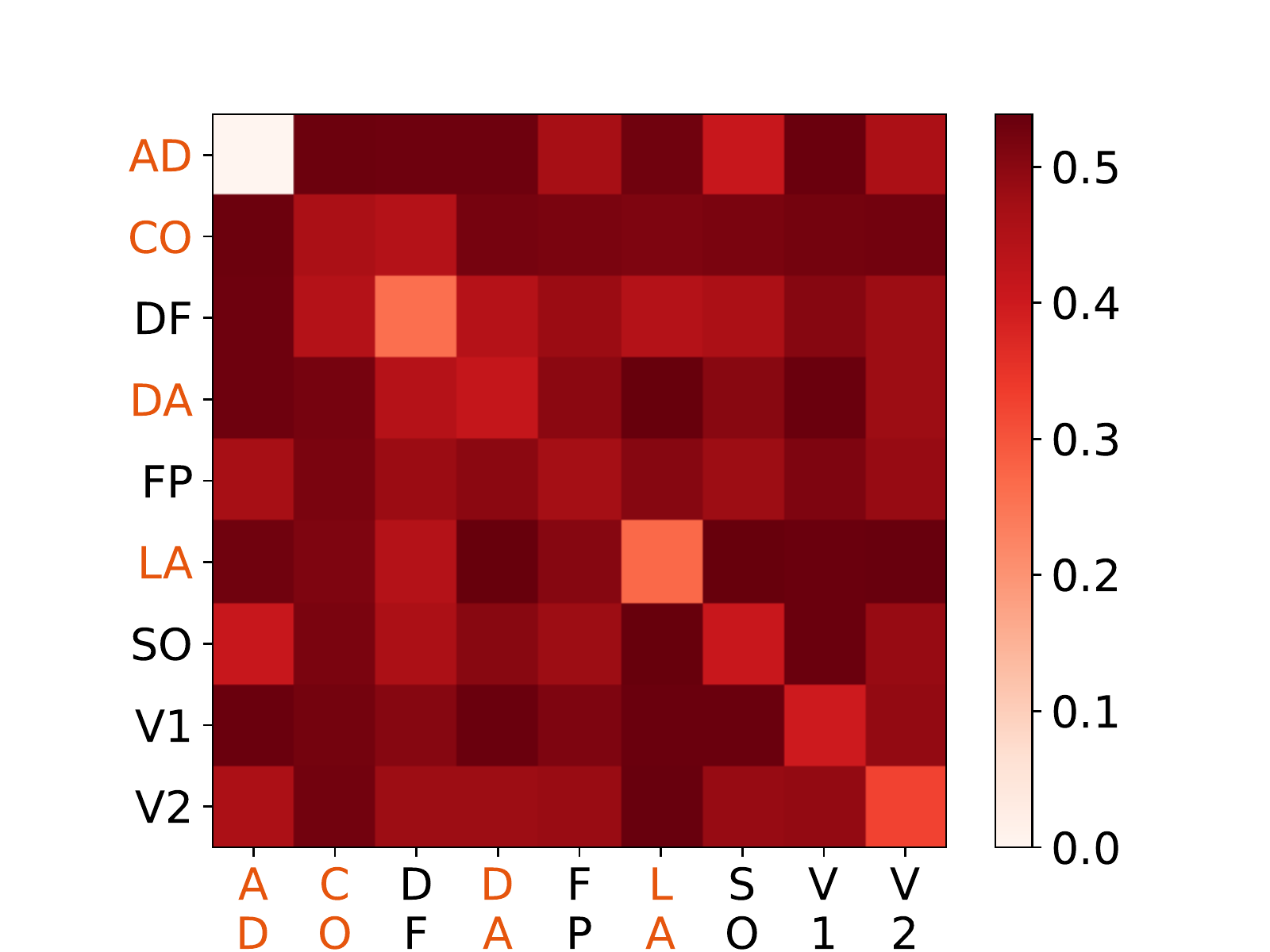}
\endminipage
\minipage{0.33\textwidth}
    \includegraphics[width=\columnwidth]{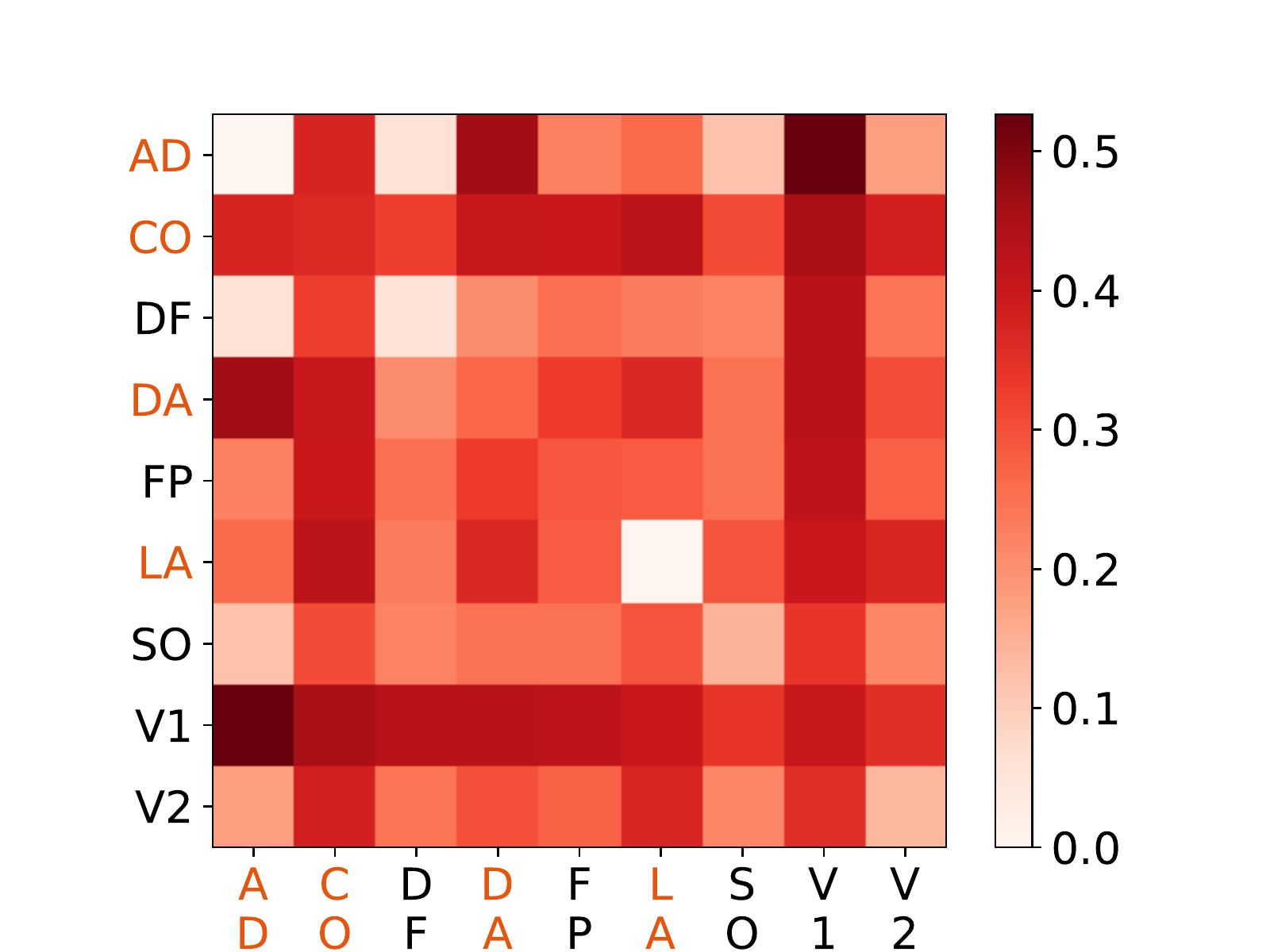}
\endminipage
\minipage{0.33\textwidth}
    \includegraphics[width=\columnwidth]{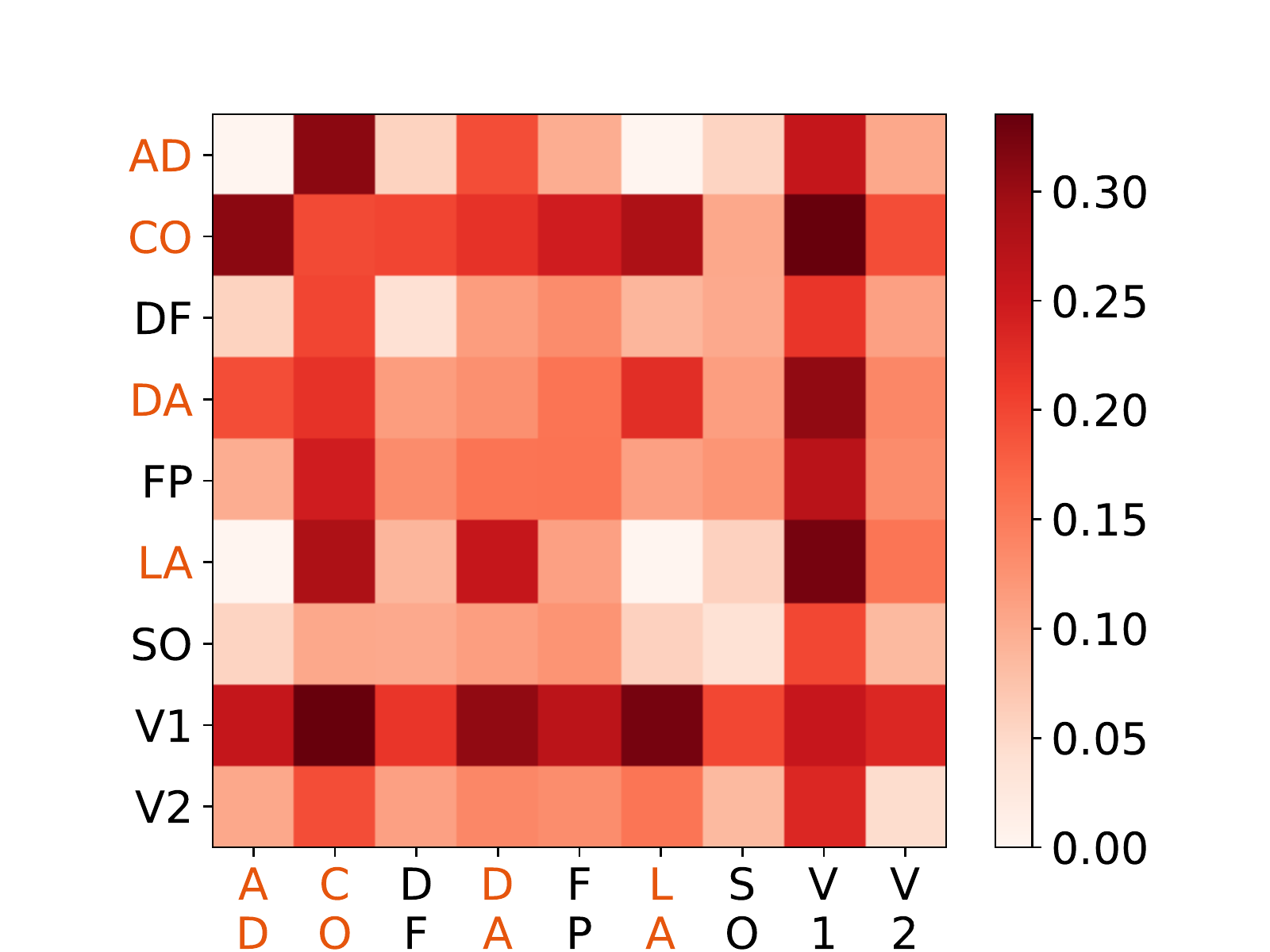}
\endminipage
    \caption{Weighted brain network edge masks at both node (top row) and subnetwork level (bottom row) for the ReadEng task, following the same setup in \Cref{exp:interpretability}.}
    \label{fig:ReadEng_interpretation}
\end{figure*}

In \Cref{fig:ReadEng_interpretation}, we provide the interpretability analysis for the ReadEng task, following the same setting as \Cref{fig:PicVocab_interpretation}. The ``ReadEng'' task involves the subjects reading aloud words presented on a screen. As can be seen in \Cref{fig:ReadEng_interpretation}, the \texttt{IGS} model effectively identifies the significance of interactions between the visual (Vis1) network and the Cingulo-Opercular (CO), Language (LA), and Dorsal Attention (DA) networks for this prediction task. Furthermore, it elucidates that the functional synchrony between the language-related networks (CO-LA, CO-AD) is accountable for this task.

% The \texttt{IGS} model finds edge mask connections between the V1 network and the CO, Language (LA), and DA networks, as well as CO-LA and CO-AD connections, to be most predictive of ``ReadEng'' performance. The ``ReadEng'' task involves the subject reading aloud words presented on a screen. From our results, it follows that the ability of Vis1 to integrate with networks responsible for language processing (LA and CO) and attention (DA), as well as the capacity for functional synchrony between the language-related networks (CO-LA), would be predictive of overall ``ReadEng'' performance. The importance of the additional CO-AD connectivity identified by our model also suggests that the ability of the CO language network to integrate with auditory centers may be involved in the neural processes responsible for proper pronunciation of the words given by visual cue. 

\end{document}